\documentclass{article}

\usepackage[utf8]{inputenc} 
\usepackage[T1]{fontenc}    

\usepackage{times}
\usepackage{graphicx} 
\usepackage{subfigure} 
\usepackage{framed}
\usepackage{natbib}

\usepackage{algorithm}
\usepackage{algorithmic}
\usepackage{todonotes}
\usepackage{wrapfig}
\usepackage{xfrac}

\usepackage[fleqn]{amsmath}
\usepackage{amsthm}
\usepackage{amssymb}

\newcommand{\af}[1]{\textcolor{black}{#1}}

\newcommand{\da}[1]{\textcolor{black}{#1}}

\usepackage{hyperref}

\usepackage[all]{hypcap} 

\usepackage{enumitem}
\usepackage[accepted]{icml2017}

\icmltitlerunning{A Closer Look at Memorization in Deep Networks}

\begin{document} 

\twocolumn[
\icmltitle{A Closer Look at Memorization in Deep Networks}
\icmlsetsymbol{equal}{*}

\begin{icmlauthorlist}
\icmlauthor{Devansh Arpit}{equal,mila,udem}
\icmlauthor{Stanisław Jastrzębski}{equal,jag}
\icmlauthor{Nicolas Ballas}{equal,mila,udem}
\icmlauthor{David Krueger}{equal,mila,udem}
\icmlauthor{Emmanuel Bengio}{mcgill}
\icmlauthor{Maxinder S. Kanwal}{ucb}
\icmlauthor{Tegan Maharaj}{mila,poly}
\icmlauthor{Asja Fischer}{bonn}
\icmlauthor{Aaron Courville}{mila,udem,cifar1}
\icmlauthor{Yoshua Bengio}{mila,udem,cifar}
\icmlauthor{Simon Lacoste-Julien}{mila,udem}
\end{icmlauthorlist}

\icmlaffiliation{mila}{Montr\'eal Institute for Learning Algorithms, Canada}
\icmlaffiliation{udem}{Universit\'e de Montr\'eal, Canada}
\icmlaffiliation{jag}{Jagiellonian University, Krakow, Poland}
\icmlaffiliation{poly}{Polytechnique Montr\'eal, Canada}
\icmlaffiliation{bonn}{University of Bonn, Bonn, Germany}
\icmlaffiliation{cifar1}{CIFAR Fellow}
\icmlaffiliation{cifar}{CIFAR Senior Fellow}
\icmlaffiliation{ucb}{University of California, Berkeley, USA}
\icmlaffiliation{mcgill}{McGill University, Canada}

\icmlcorrespondingauthor{}{david.krueger@umontreal.ca}

\icmlkeywords{deep learning, deep networks, capacity, regularization}

\vskip 0.3in
]

\printAffiliationsAndNotice{\icmlEqualContribution} 

\begin{abstract}
We examine the role of memorization in deep learning, drawing connections to capacity, generalization, and adversarial robustness.
While deep networks are capable of memorizing noise data, our results suggest that they tend to prioritize learning simple patterns first.
In our experiments, we expose qualitative differences in gradient-based optimization of deep neural networks (DNNs) on noise vs.~real data.
We also demonstrate that for appropriately tuned explicit regularization (e.g.,~dropout) we can degrade DNN training performance on noise datasets without compromising generalization on real data.
Our analysis suggests that the notions of effective capacity which are dataset independent are unlikely to explain the generalization performance of deep networks when trained with gradient based methods because training data itself plays an important role in determining the degree of memorization.
\end{abstract}

\section{Introduction}

The traditional view of generalization holds that a model with sufficient capacity (e.g. more parameters than training examples) will be able to ``memorize'' each example, overfitting the training set and yielding poor generalization to validation and test sets \citep{dl_book}.
Yet deep neural networks (DNNs) often achieve excellent generalization performance with massively over-parameterized models. This phenomenon is not well-understood.

From a representation learning perspective, the generalization capabilities of DNNs \da{are believed to} stem from their incorporation of good generic priors (see, e.g., \citet{bengio2009}). 
\citet{Tegmark} further suggest that the priors of deep learning are well suited to the physical world. 
But while the priors of deep learning may help explain why DNNs learn to efficiently represent complex real-world functions, they are not restrictive enough to rule out memorization.

On the contrary, deep nets are known to be universal approximators, capable of representing arbitrarily complex functions given sufficient capacity \cite{cybenko1989approximation, hornik1989multilayer}.  
Furthermore, recent work has shown that the expressiveness of DNNs grows exponentially with depth \cite{montufar2014,ganguli2016}.
These works, however, only examine the \textit{representational capacity}, that is, the set of hypotheses a model is capable of expressing via some value of its parameters.

Because DNN optimization is not well-understood, it is unclear which of these hypotheses can actually be reached by gradient-based training \cite{bottou1998online}.
In this sense, optimization and generalization are entwined in DNNs.
To account for this, we formalize a notion of the \textit{effective capacity (EC)} of a learning algorithm $\mathcal{A}$ (defined by specifying both the model {\it and the training procedure}, e.g.,``train the LeNet architecture \citep{lecun1998mnist} for 100 epochs using \af{stochastic gradient descent} (SGD) with a learning rate of $0.01$'') as the set of hypotheses which can be reached by applying that learning algorithm on {\it some} dataset.  Formally, using set-builder notation:
$$
EC(\mathcal{A}) = \{h \mid  \exists \mathcal{D} \text{ such that } h \in \mathcal{A}(\mathcal{D}) \} \enspace,
$$
where $\mathcal{A}(\mathcal{D})$
represents the set of hypotheses that is reachable by $\mathcal{A}$ on a dataset $\mathcal{D}$\footnote{
Since $\mathcal{A}$ can be stochastic, $\mathcal{A}(\mathcal{D})$ is a set.
}.

One might suspect that DNNs effective capacity is sufficiently limited by gradient-based training and early stopping to resolve the apparent paradox between DNNs' excellent generalization and their high representational capacity.
However, the experiments of \citet{understanding_DL} suggest that this is not the case.
They demonstrate that DNNs are able to fit pure noise without even needing substantially longer training time. 
Thus even the {\it effective} capacity of DNNs may be too large, from the point of view of traditional learning theory.

By demonstrating the ability of DNNs to ``memorize'' random noise, \citet{understanding_DL} also raise the question whether deep networks use similar memorization tactics on real datasets.
Intuitively, a brute-force memorization approach to fitting data does not capitalize on patterns shared between training examples or features; the \emph{content} of what is memorized is irrelevant.
A paradigmatic example of a memorization algorithm is k-nearest neighbors \citep{fix1951discriminatory}.
Like \citet{understanding_DL}, we do not formally define memorization; rather, we investigate this intuitive notion of memorization by training DNNs to fit random data.

\subsection*{Main Contributions}
 We operationalize the definition of ``memorization'' as {\it the behavior exhibited by DNNs trained on noise}, and conduct a series of experiments that contrast the learning dynamics of DNNs on real vs.~noise data.
 Thus, our analysis builds on the work of \citet{understanding_DL} and further investigates the role of memorization in DNNs. 

Our findings are summarized as follows:
\begin{enumerate}
\item There are qualitative differences in DNN optimization behavior on real data vs.~noise.
In other words, DNNs do not just memorize real data (Section ~\ref{sec:qualitative_differences}).
\item DNNs learn simple patterns first, before memorizing (Section ~\ref{sec:simple_patterns_first}).
In other words, DNN optimization is {\it content-aware}, taking advantage of patterns shared by multiple training examples.
\item Regularization techniques can differentially hinder memorization in DNNs while preserving their ability to learn about real data (Section ~\ref{sec:regularization}). 
\end{enumerate}

\section{Experiment Details}

We perform experiments on MNIST \citep{lecun1998mnist} and CIFAR10 \citep{CIFAR10} datasets.
We investigate two classes of models: 2-layer multi-layer perceptrons (MLPs) with rectifier linear units (ReLUs) on MNIST and convolutional neural networks (CNNs) on CIFAR10.
If not stated otherwise, the MLPs have 4096 hidden units per layer and are trained for $1000$ epochs with SGD and learning rate $0.01$. 
The CNNs are a small Alexnet-style CNN\footnote{Input $\rightarrow$ Crop(2,2) $\rightarrow$ Conv(200,5,5) $\rightarrow$ BN $\rightarrow$ ReLU $\rightarrow$ MaxPooling(3,3) $\rightarrow$ Conv(200,5,5) $\rightarrow$ BN$\rightarrow$ ReLU$\rightarrow$ MaxPooling(3,3) $\rightarrow$ Dense(384) $\rightarrow$ BN $\rightarrow$ ReLU $\rightarrow$ Dense(192) $\rightarrow$ BN $\rightarrow$ ReLU $\rightarrow$ Dense($\#$classes) $\rightarrow$ Softmax. Here Crop(. , .) crops height and width from both sides with respective values. } (as in \citet{understanding_DL}), and are trained using SGD with momentum=$0.9$ and learning rate of $0.01$, scheduled to drop by half every 15 epochs.

Following \citet{understanding_DL}, in many of our experiments we replace either (some portion of) the labels (with random labels), or the inputs (with i.i.d.~Gaussian noise matching the real dataset's mean and variance) for some fraction of the training set. 
We use \emph{randX} and \emph{randY} to denote datasets with (100\%, unless specified) noisy inputs and labels (respectively).

\section{Qualitative Differences of DNNs Trained on Random vs. Real Data}
\label{sec:qualitative_differences}

\citet{understanding_DL} empirically demonstrated that DNNs are capable of fitting random data, which implicitly necessitates some high degree of memorization.
In this section, we investigate whether DNNs employ similar memorization strategy when trained on real data.  
In particular, our experiments highlight some qualitative differences between DNNs trained on real data vs.~random data, supporting the fact that DNNs do not use brute-force memorization to fit real datasets.

\subsection{Easy Examples as Evidence of Patterns in Real Data}
\label{sec:easy_examples}
\begin{figure}[!t]
  \center
  \label{fig:easy_hard.pdf}
  \includegraphics[width=7.5cm]{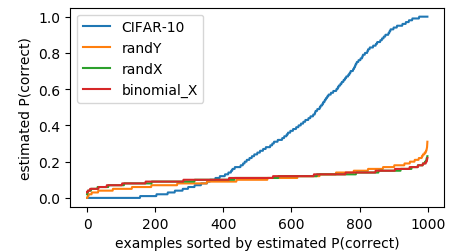}
  \caption{Average (over 100 experiments) misclassification rate for each of 1000 examples after one epoch of training.
           This measure of an example's difficulty is much more variable in real data.
       We conjecture this is because the easier examples are explained by some simple patterns, which are reliably learned within the first epoch of training.
	   We include 1000 points samples from a binomial distribution with $n=100$ and $p$ equal to the average estimated P(correct) for randX, and note that this curve closely resembles the randX curve, suggesting that random inputs are all equally difficult.
}
\end{figure}

A brute-force memorization approach to fitting data should apply equally well to different training examples.
However, if a network is learning based on patterns in the data, some examples may fit these patterns better than others.
We show that such ``easy examples'' (as well as correspondingly ``hard examples'') are common in real, but not in random, datasets.
Specifically, for each setting (real data, randX, randY), we train an MLP for a single epoch starting from 100 different random initializations and shufflings of the data.
We find that, for real data, many examples are consistently classified (in)correctly after a single epoch, suggesting that different examples are significantly easier or harder in this sense.
For noise data, the difference between examples is much less, indicating that these examples are fit (more) independently.
Results are presented in Figure~\ref{fig:easy_hard.pdf}.

\begin{figure}[!t]
  \center
  \label{fig:filters.pdf}
  \includegraphics[width=7.5cm]{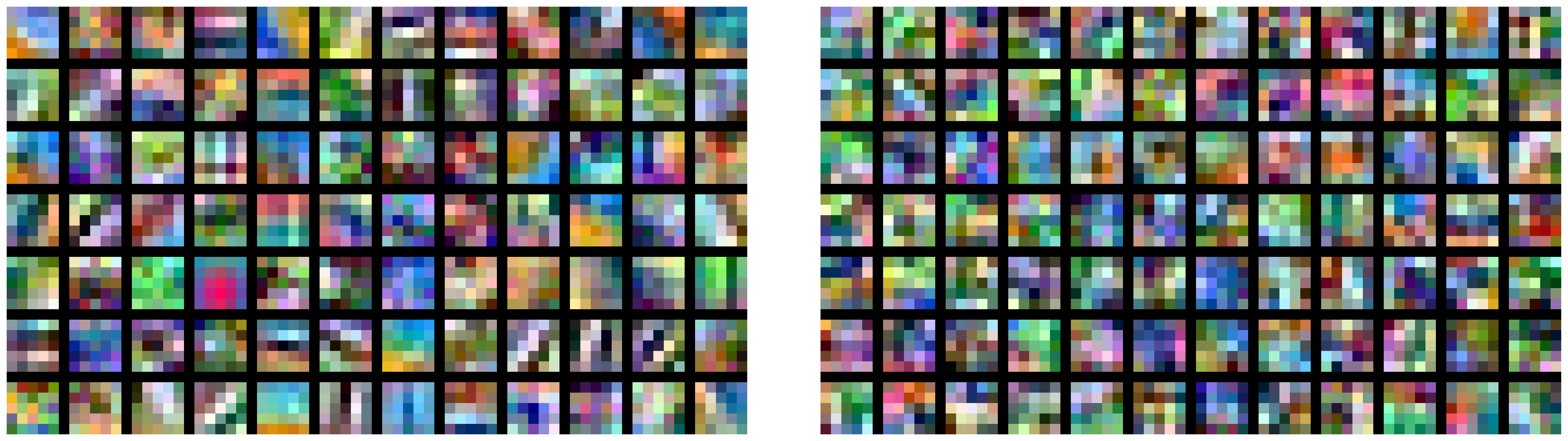}
  \caption{Filters from first layer of network trained on CIFAR10 (left) and randY (right).}
\end{figure}

For randX, apparent differences in difficulty are well modeled as random Binomial noise.
For randY, this is not the case, indicating some use of shared patterns. 
Visualizing first-level features learned by a CNN supports this hypothesis (Figure~\ref{fig:filters.pdf}).

\begin{figure*}[!ht]
  \centering
  \includegraphics[width=6.1cm]{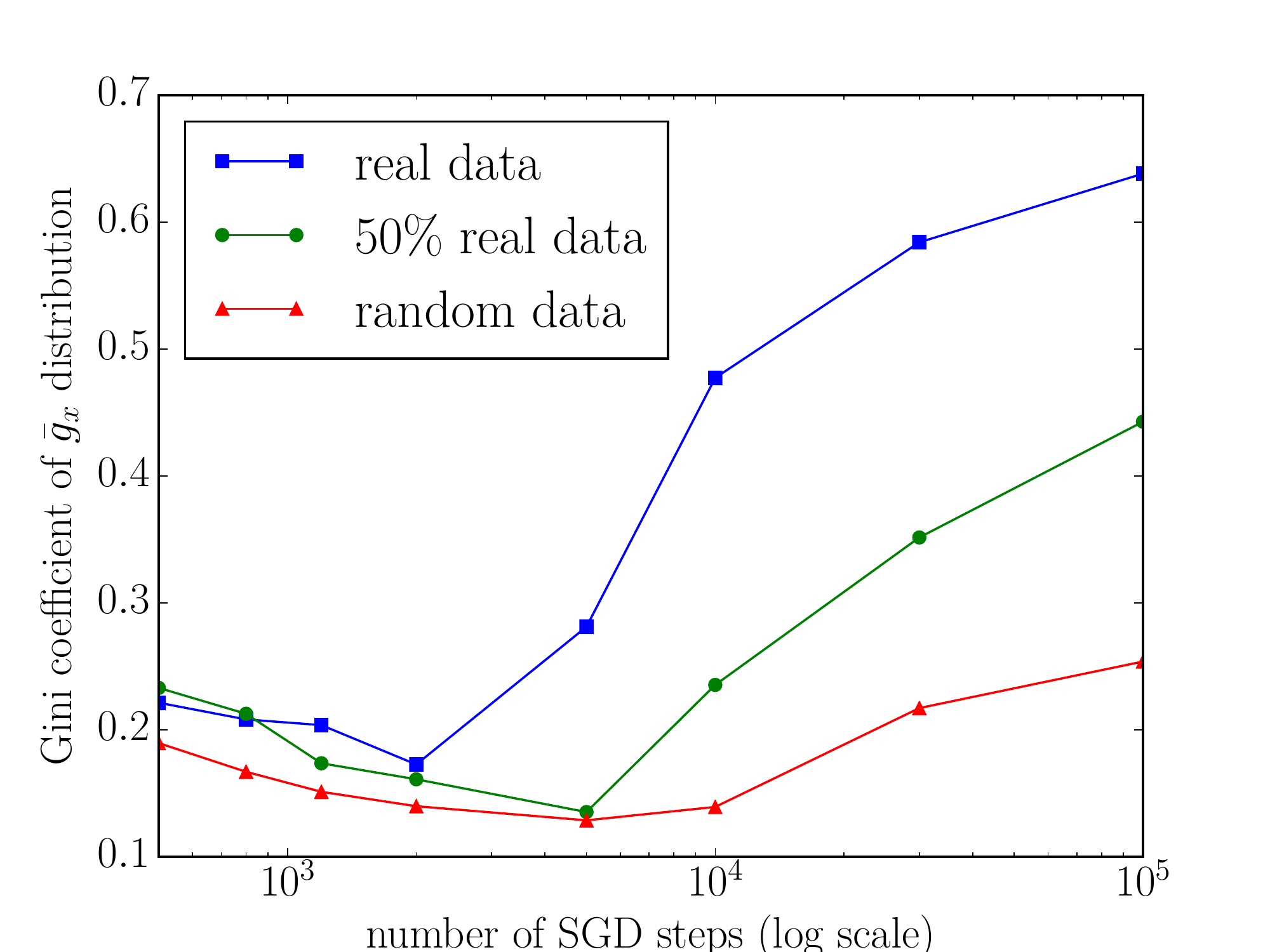}
  \hspace{-1em}
  \includegraphics[width=6.1cm]{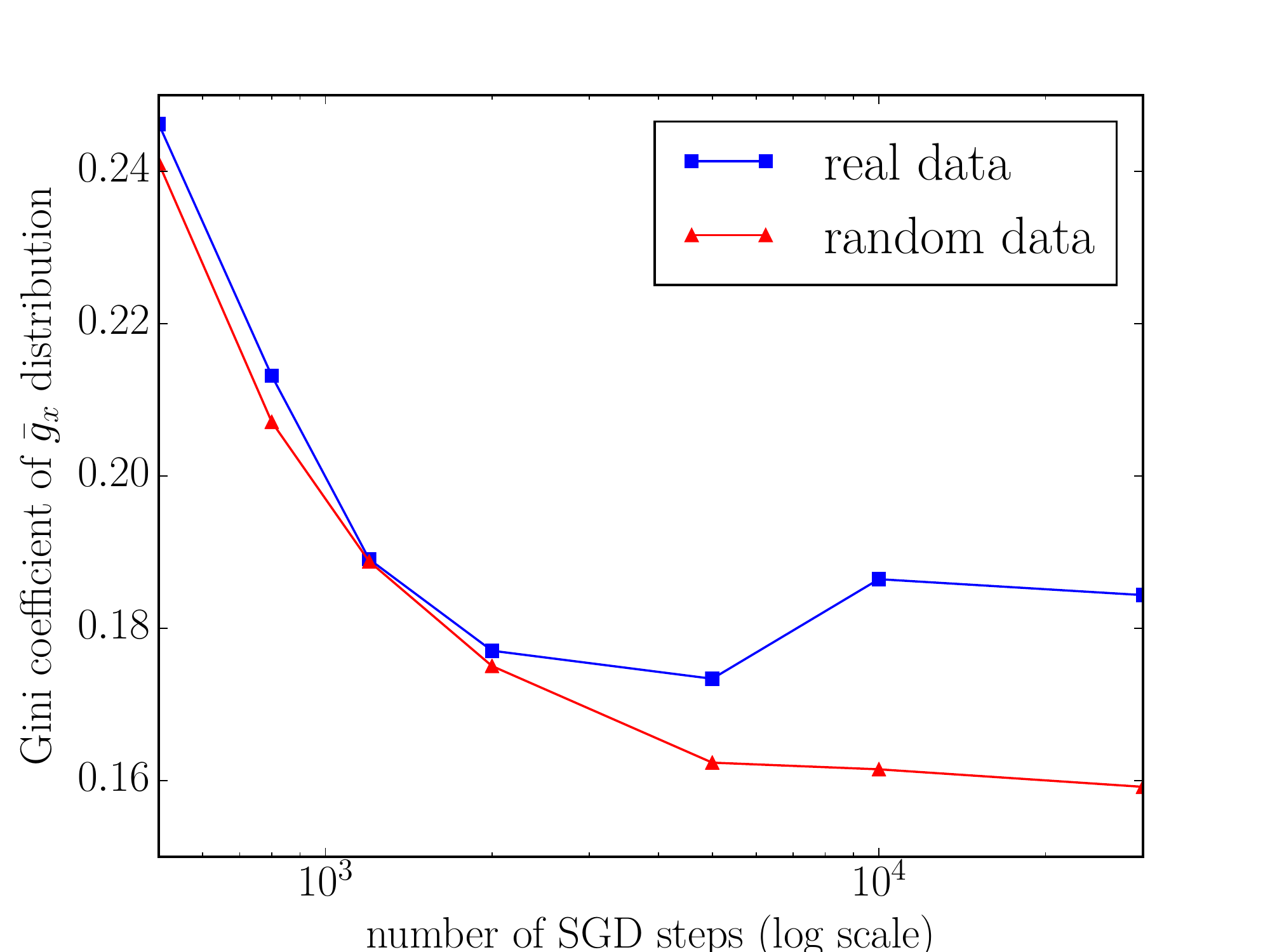}
  \caption{Plots of the Gini coefficient of $\bar{g}_\mathbf{x}$ over examples $\mathbf{x}$ (see section \ref{sec:grad_sens}) as training progresses, for a 1000-example real dataset (14x14 MNIST) versus random data. On the left, $Y$ is the normal class label; on the right, there are as many classes as examples, the network has to learn to map each example to a unique class.
    }
  \label{fig:grad_sensitivity}
\end{figure*}

\subsection{Loss-Sensitivity in Real vs. Random Data}
\label{sec:grad_sens}

To further investigate the difference between real and fully random inputs, we propose a proxy measure of memorization via gradients. Since we cannot measure quantitatively how much each training sample $\mathbf{x}$ is memorized, we instead measure the effect of each sample on the average loss. 
That is, we measure the norm of the loss gradient with respect to a previous example $\mathbf{x}$ after $t$ SGD updates. 
Let $\mathcal{L}_t$ be the loss after $t$ updates; then the sensitivity measure is given by
$$g^t_\mathbf{x}=\|\partial \mathcal{L}_t/\partial \mathbf{x}\|_1 \enspace.$$
The parameter update from training on $\mathbf{x}$ influences all future $\mathcal{L}_t$ indirectly by changing the subsequent updates on different training examples.
We denote the average over $g^t_\mathbf{x}$ after $T$ steps as $\bar{g}_\mathbf{x}$, and refer to it as \emph{loss-sensitivity}. Note that we only report $\ell^1$-norm results, but that results stay very similar using  $\ell^2$-norm and infinity norm. 

We compute $g^t_\mathbf{x}$ by unrolling $t$ SGD steps and applying backpropagation over the unrolled computation graph, as done by \citet{maclaurin2015gradient}.
Unlike \citet{maclaurin2015gradient}, we only use this procedure to compute $g^t_\mathbf{x}$, and do not modify the training procedure in any way.

We find that for real data, only a subset of the training set has high $\bar{g}_\mathbf{x}$, while for random data, $\bar{g}_\mathbf{x}$ is high for virtually all examples. We also find a different behavior when \textit{each example} is given a unique class; in this scenario, the network has to learn to identify each example uniquely,
yet still behaves differently when given real data than when given random data as input.

We visualize  (Figure \ref{fig:grad_sensitivity}) the spread of $\bar{g}_\mathbf{x}$ as training progresses by computing the Gini coefficient over $\mathbf{x}$'s. The Gini coefficient \citep{gini} is a measure of the inequality among values of a frequency distribution; a coefficient of 0 means \af{exact equality (i.e.,~all values are the same), while a coefficient of 1 means maximal inequality among values.}
We observe that, when trained on real data, the network has a high $\bar{g}_\mathbf{x}$ for a few examples, while on random data the network is sensitive to most examples. The difference between the random data scenario, where we know the neural network needs to do memorization, and the real data scenario, where we're trying to understand what happens, leads us to believe that this measure is indeed sensitive to memorization. Additionally, these results suggest that when being trained on real data, the neural network probably does not memorize, or at least not in the same manner it needs to for random data.

\begin{figure}[h]
  \centering
  \hspace{-2.0em}
  \includegraphics[width=8.6cm]{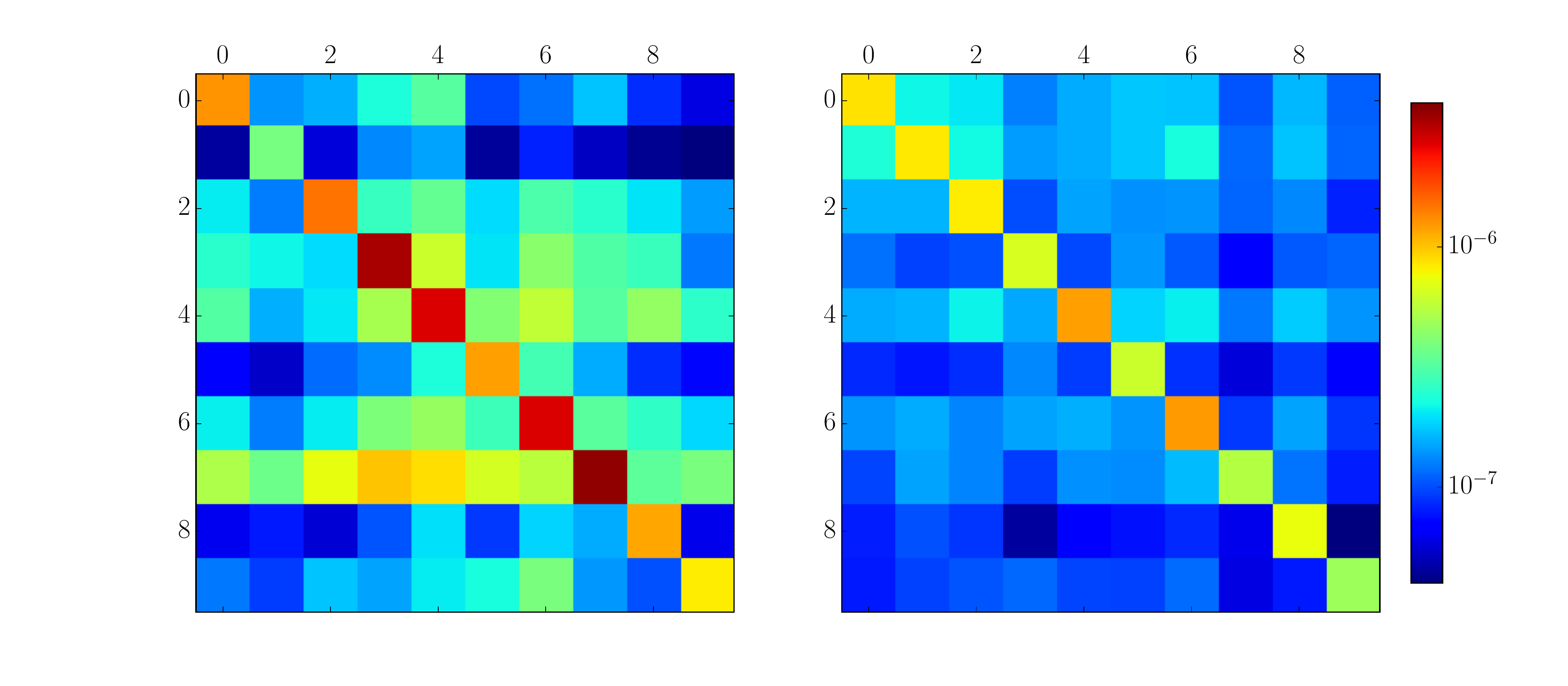}
  \caption{Plots of per-class $g_x$ (see previous figure; log scale), a cell $i,j$ represents the average $|\partial \mathcal{L}(y=i)/\partial x_{y=j}|$, i.e. the loss-sensitivity of examples of class $i$ w.r.t. training examples of class $j$. Left is real data, right is random data.}
  \label{fig:grad_class_sensitivity}
\end{figure}

In addition to the different behaviors for real and random data described above, we also consider a class specific loss-sensitivity: $\bar{g}_{i, j} =\mathbb{E}_{(x,y)}\sfrac{1}{T}\sum_t^T|\partial \mathcal{L}_t(y=i)/\partial x_{y=j}|$, where $\mathcal{L}_t(y=i)$ is the term in the crossentropy sum corresponding to class $i$.
We observe that the loss-sensitivity 
\af{w.r.t.~class $i$ for}
training examples of class $j$
is higher when $i=j$, but more spread out for real data (see Figure \ref{fig:grad_class_sensitivity}).
An interpretation of this is that for real data there are more interesting cross-category patterns that can be learned than for random data.

Figure \ref{fig:grad_sensitivity} and \ref{fig:grad_class_sensitivity} were obtained by training a fully-connected network with 2 layers of 16 units on 1000 downscaled $14\times 14$ MNIST digits using SGD.

\subsection{Capacity and Effective Capacity}

In this section, we investigate the impact of capacity and effective capacity on learning of datasets having different amounts of random input data or random labels.

\subsubsection{Effects of capacity and dataset size on validation performances}

\begin{figure}[!ht]
\centering
\includegraphics[scale=.072]{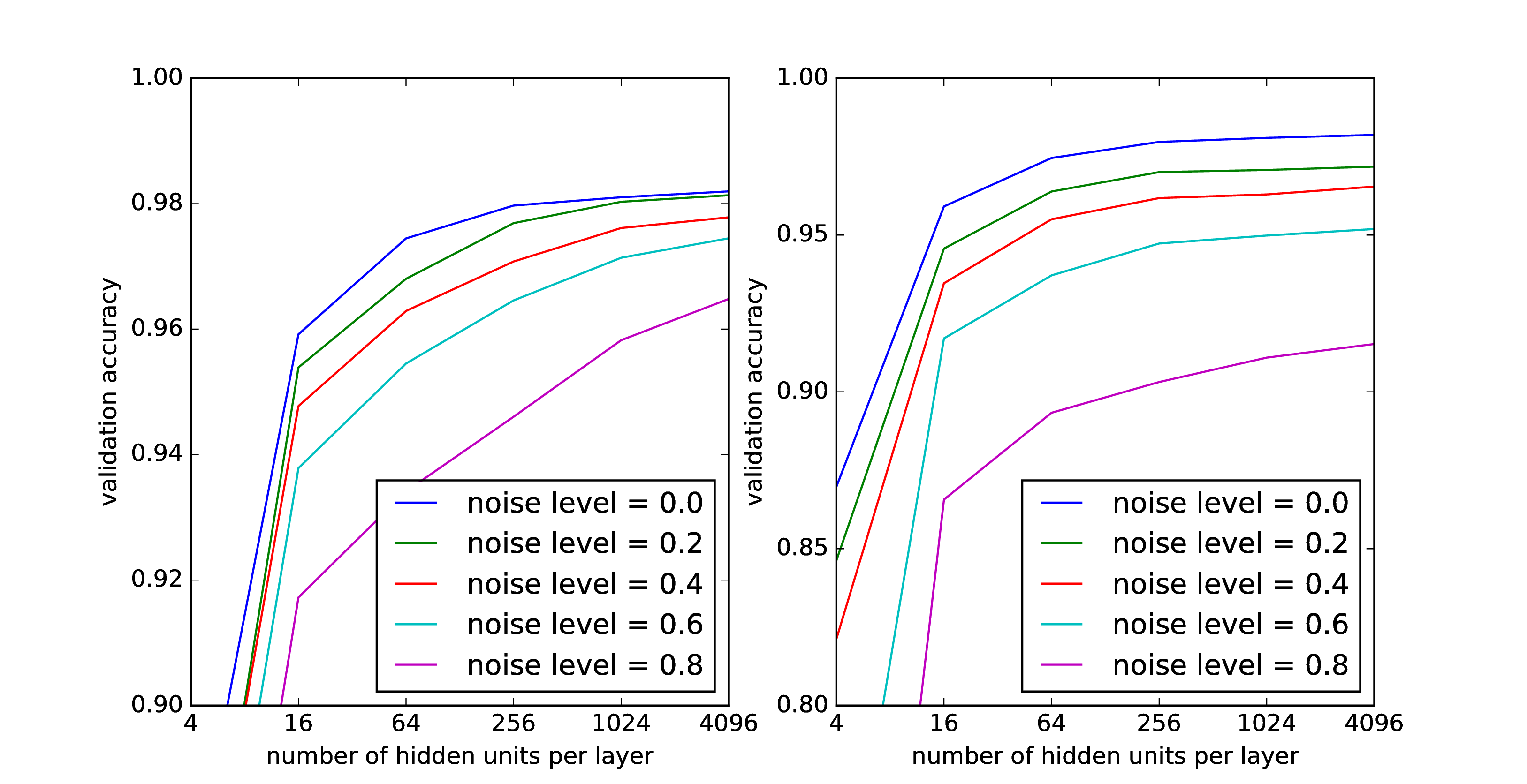}
\caption{Performance as a function of capacity in 2-layer MLPs \af{trained on (noisy versions of) MNIST}. For real data, performance is already very close to maximal with 4096 hidden units, but when there is noise in the dataset, higher capacity is needed.}
\label{fig:val_acc__vs__capacity}
\end{figure}

In a first experiment, we study how overall model capacity impacts the validation performances for datasets with different amounts of noise.
On MNIST, we found that the optimal validation performance requires a higher capacity model in the presence of noise examples (see Figure~\ref{fig:val_acc__vs__capacity}).
This trend was consistent for noise inputs on CIFAR10, but we did not notice any relationship between capacity and validation performance on random \emph{labels} on CIFAR10. 

This result contradicts the intuitions of traditional learning theory, which suggest that capacity should be restricted, in order to enforce the learning of (only) the most regular patterns.
Given that DNNs can perfectly fit the training set in any case, we hypothesize that that higher capacity allows the network to fit the noise examples in a way that does not interfere with learning the real data.
In contrast, if we were simply to \emph{remove} noise examples, yielding a smaller (clean) dataset, a \emph{lower} capacity model would be able to achieve optimal performance. 
	
\begin{figure*}[!t]
   \center
  \subfigure
  {
    \includegraphics[scale=.08]{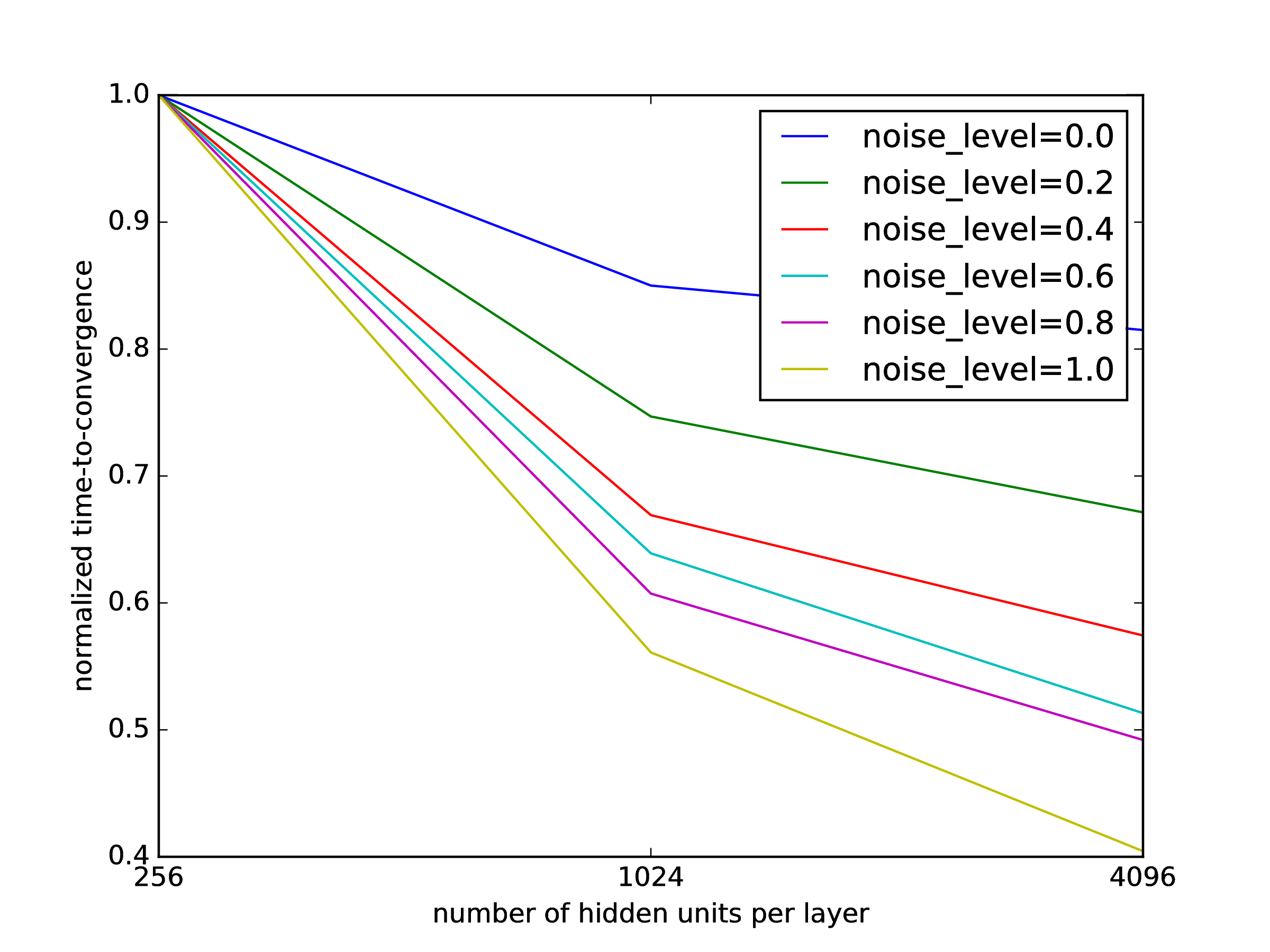}
  }
  \hspace{1mm}
  \subfigure
  {
    \includegraphics[scale=.08]{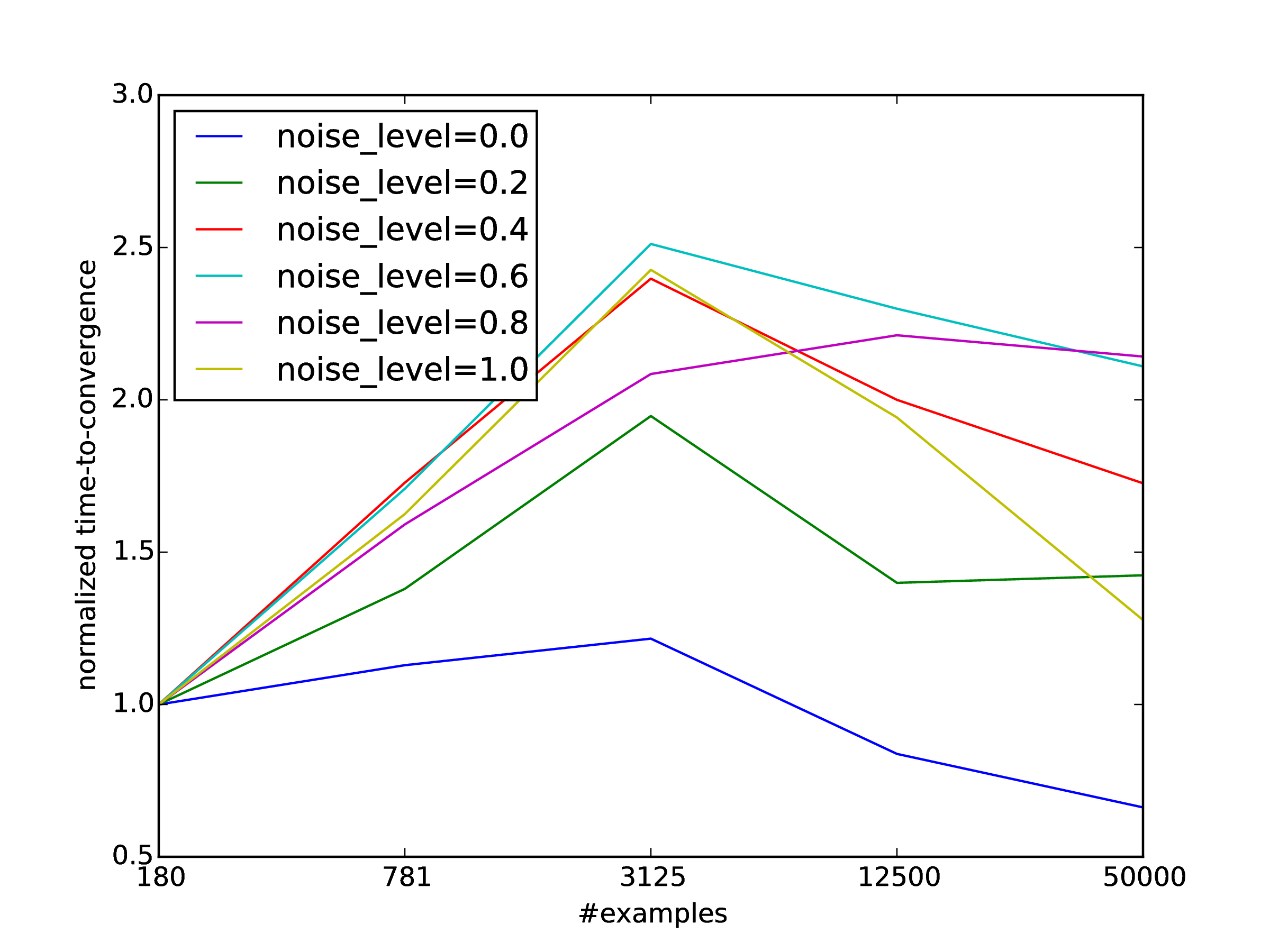}
  }
   \caption{
Time to convergence as a function of capacity with dataset size fixed to 50000 (left), or dataset size with capacity fixed to 4096 units (right).
``Noise level'' denotes to the proportion of training points whose inputs are replaced by Gaussian noise.
Because of the patterns underlying real data, having more capacity/data does not decrease/increase training time as much as it does for noise data.
}
   \label{fig:ttc__vs__capacity}
\end{figure*}

\subsubsection{Effects of capacity and dataset size on training time}
Our next experiment measures time-to-convergence, i.e. how many epochs it takes to reach 100\% training accuracy.
Reducing the capacity or increasing the size of the dataset slows down training as well for real as for noise data\footnote{
Regularization can also increase time-to-convergence; see section~\ref{sec:regularization}.
}.
However, the effect is more severe for datasets containing noise, as our experiments in this section show (see Figure~\ref{fig:ttc__vs__capacity}).

Effective capacity of a DNN can be increased by increasing the representational capacity (e.g.~adding more hidden units) or training for longer.
Thus, increasing the number of hidden units decreases the number of training iterations needed to fit the data, up to some limit.
We observe \emph{stronger} diminishing returns from increasing representational capacity for real data, indicating that this limit is lower, and a smaller representational capacity is sufficient, for real datasets.

Increasing the number of examples (keeping representational capacity fixed) also increases the time needed to memorize the training set.
In the limit, the representational capacity is simply insufficient, and memorization is not feasible.
On the other hand, when the relationship between inputs and outputs is meaningful, new examples simply give more (possibly redundant) clues as to what the input $\rightarrow$ output mapping is.
Thus, in the limit, an idealized learner should be able to predict unseen examples perfectly, absent noise.
Our experiments demonstrate that time-to-convergence is not only longer on noise data (as noted by \citet{understanding_DL}), but also, \emph{increases} substantially as a function of dataset size, relative to real data.
Following the reasoning above, this suggests that our networks are learning to extract patterns in the data, rather than memorizing.

        \begin{figure*}[!t]
		\center
		\subfigure[Noise added on classification inputs.]
		{
			\label{fig:mnist_noisey_acc}
			\includegraphics[width=4.1cm]{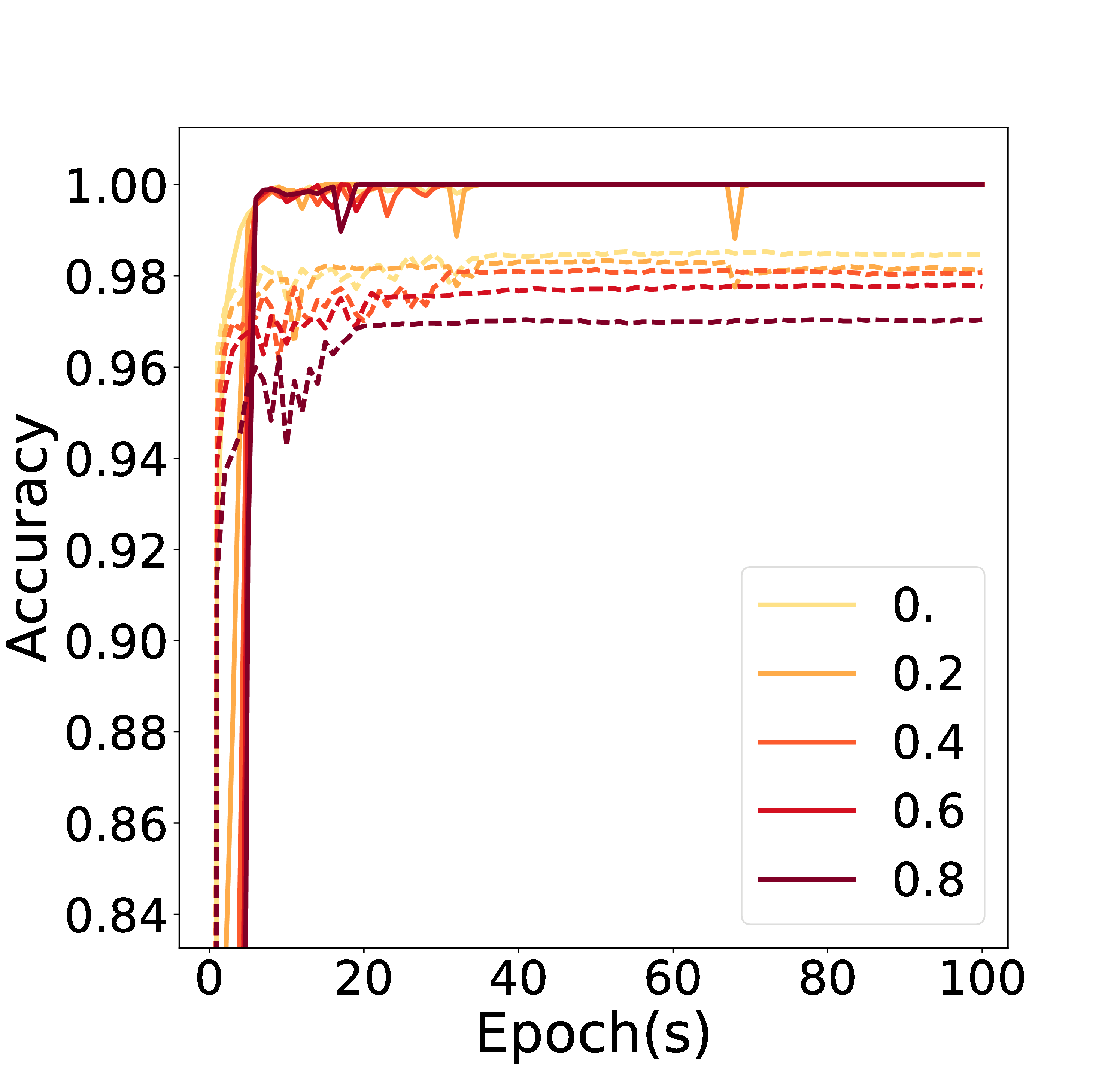}
            \includegraphics[width=4.1cm]{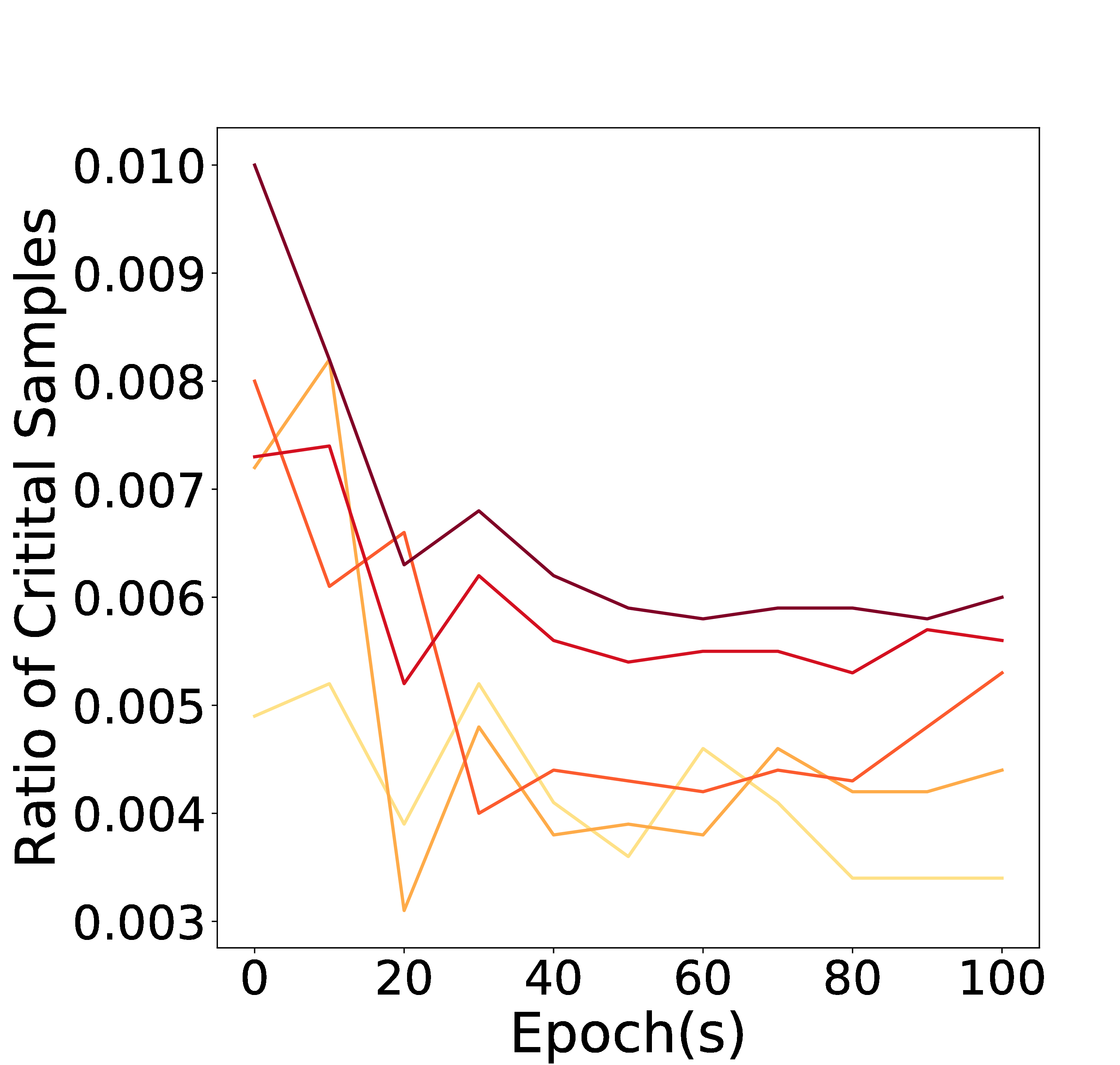}
                                                
		}
		\hspace{0.3mm}
		\subfigure[Noise added on classification labels.]
		 {
                   \includegraphics[width=4.1cm]{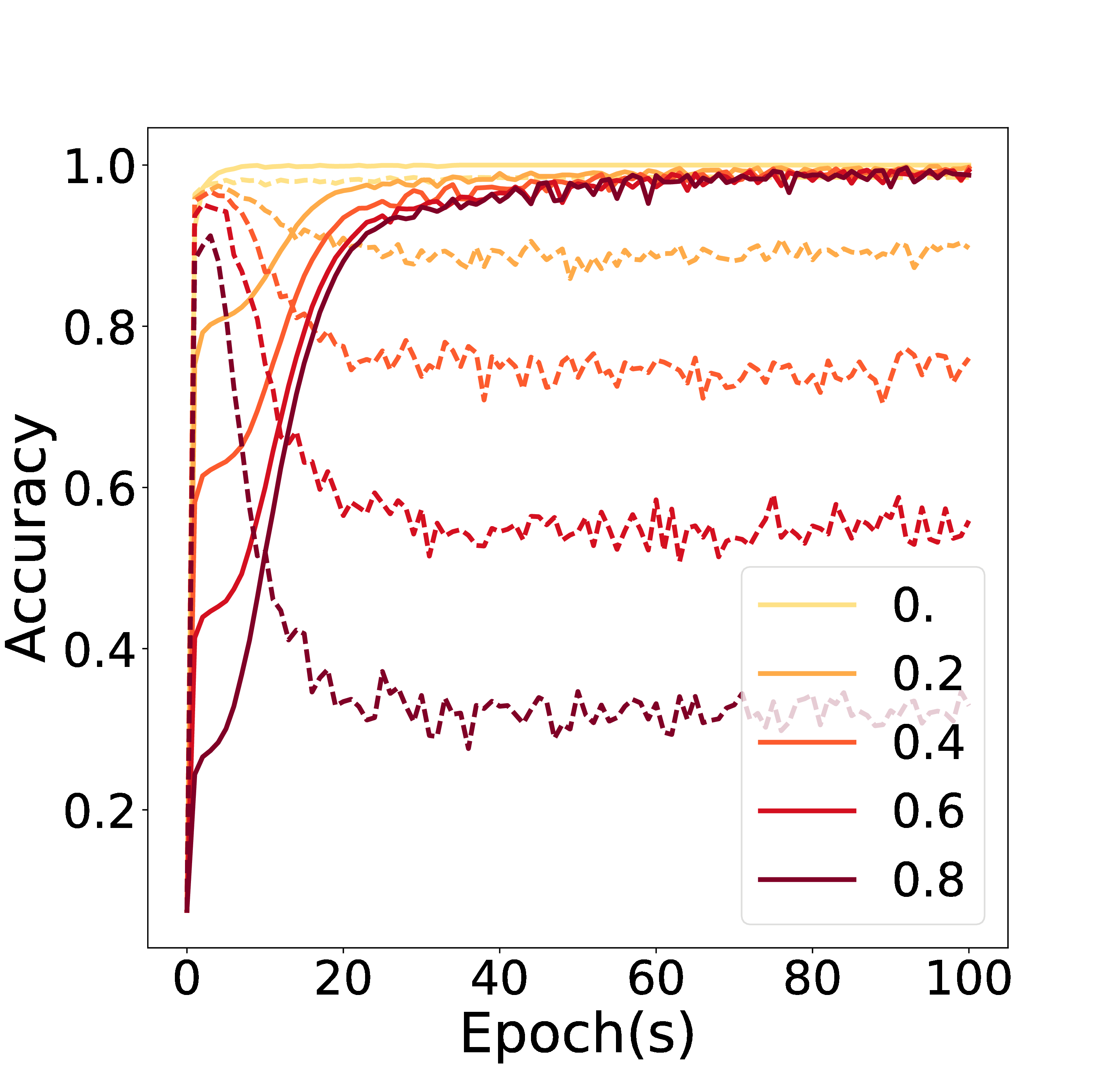}
                   \includegraphics[width=4.1cm]{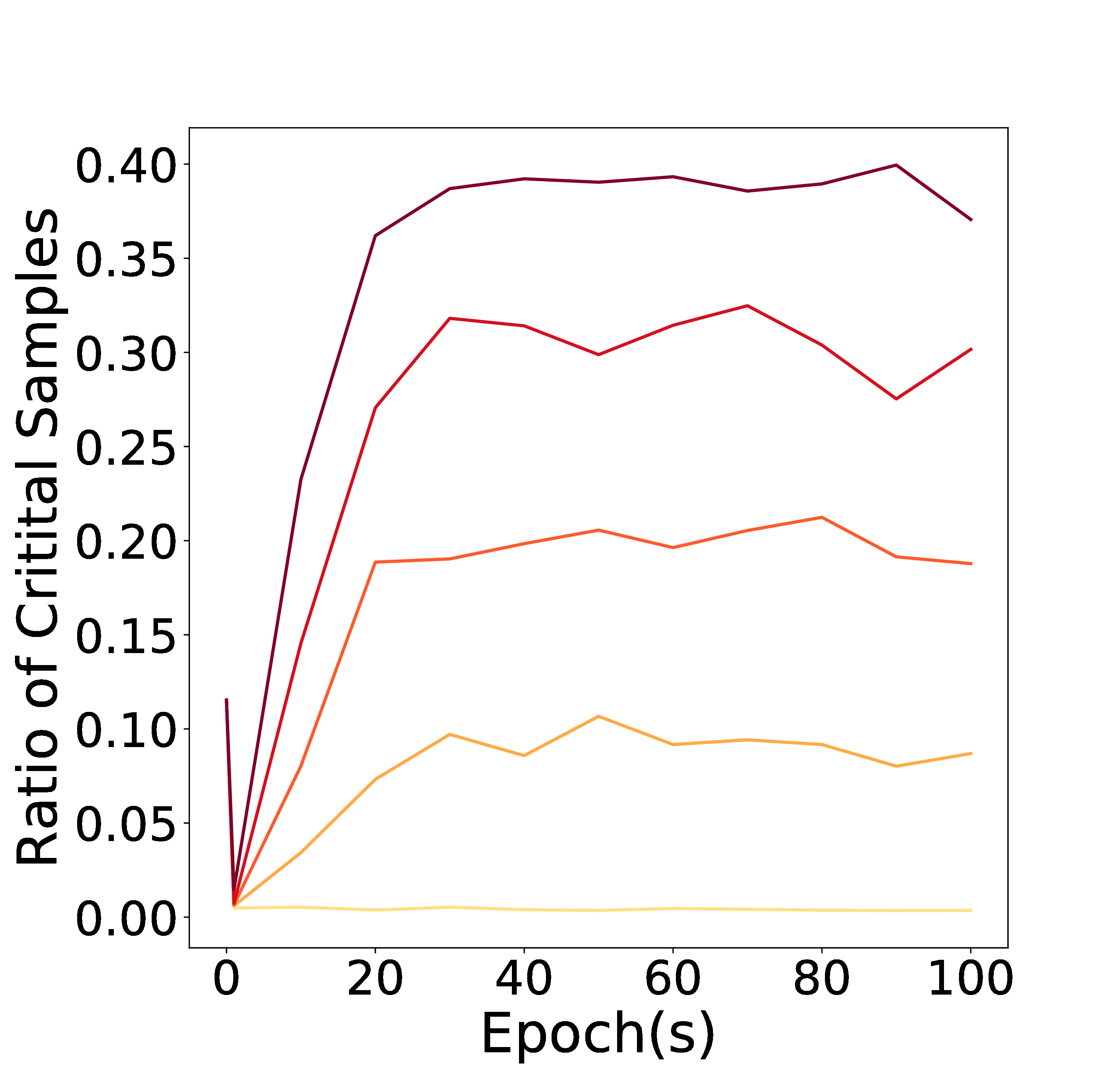}
		   \label{fig:mnist_noisey_critical}
		}
		\caption{Accuracy (left in each pair, solid is train, dotted is validation) and Critical sample ratios (right in each pair) for MNIST.}

	\label{fig:mnist_nicolas}
	\end{figure*}

        \begin{figure*}[!t]
		\center
		\subfigure[Noise added on classification inputs.]
		{
			\label{fig:cifar10_noisey_acc}
			\includegraphics[width=4.1cm]{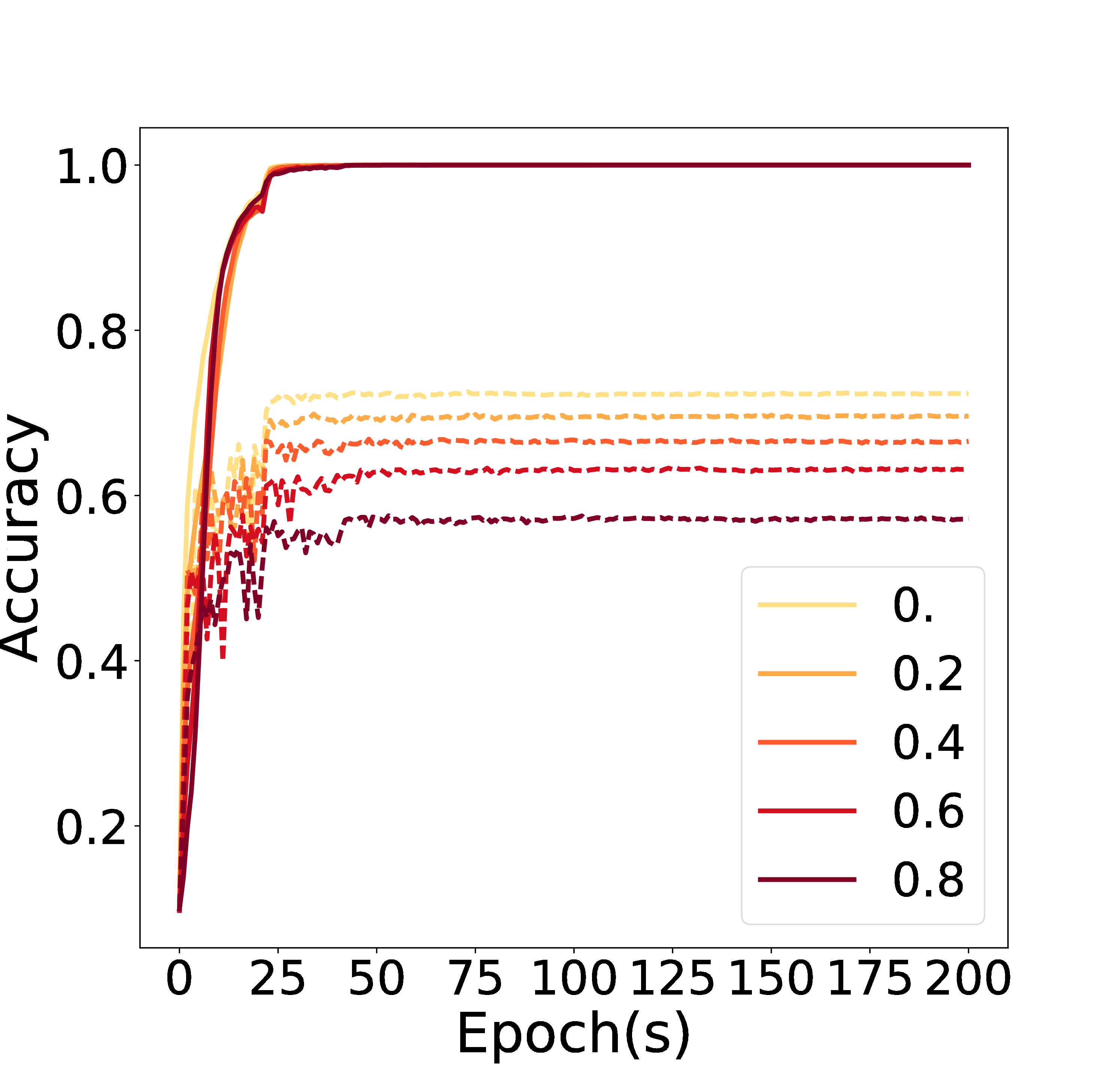}
            \includegraphics[width=4.1cm]{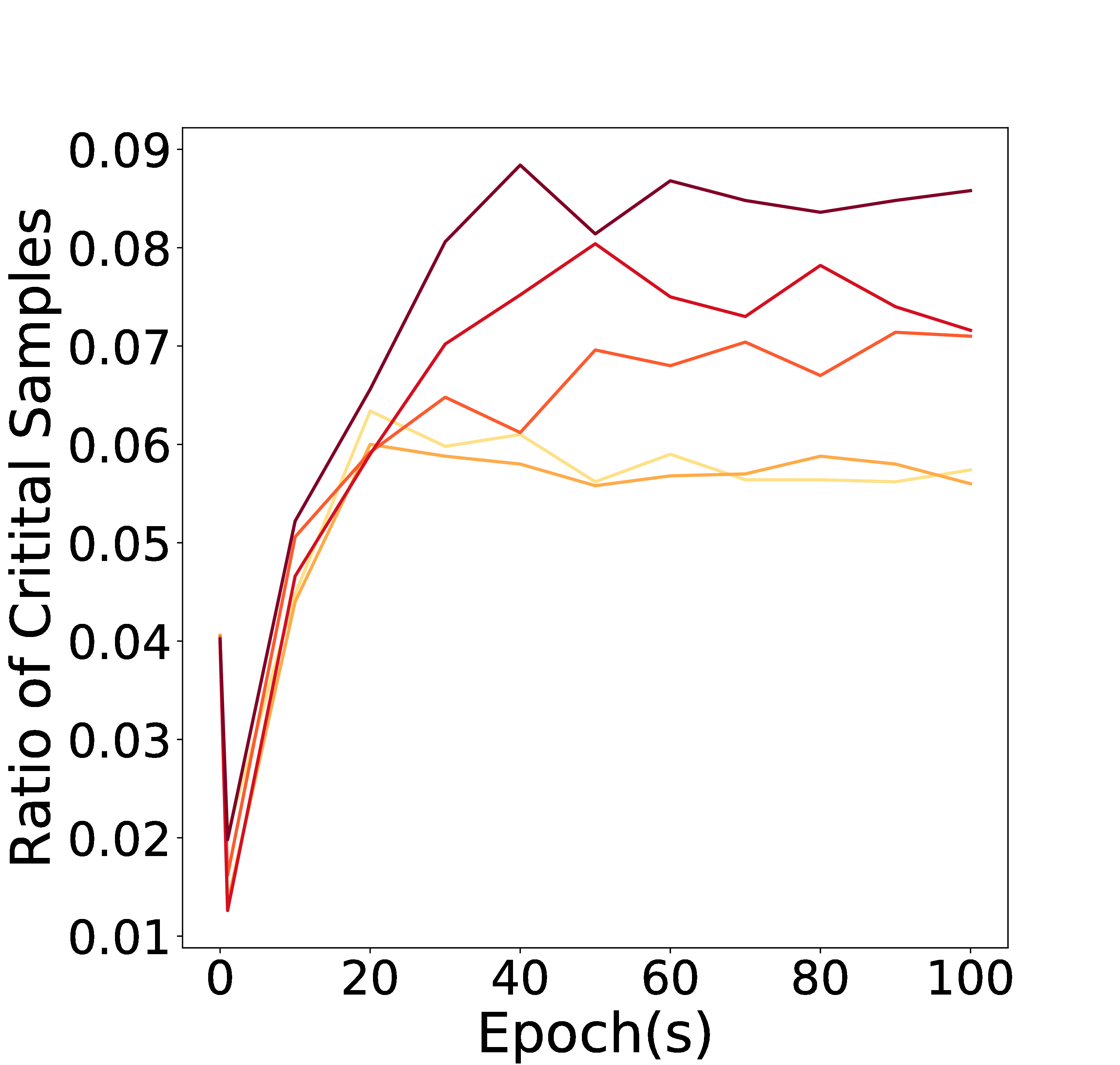}
                                                
		}
		\hspace{0.3mm}
		\subfigure[Noise added on classification labels.]
		 {
                   \includegraphics[width=4.1cm]{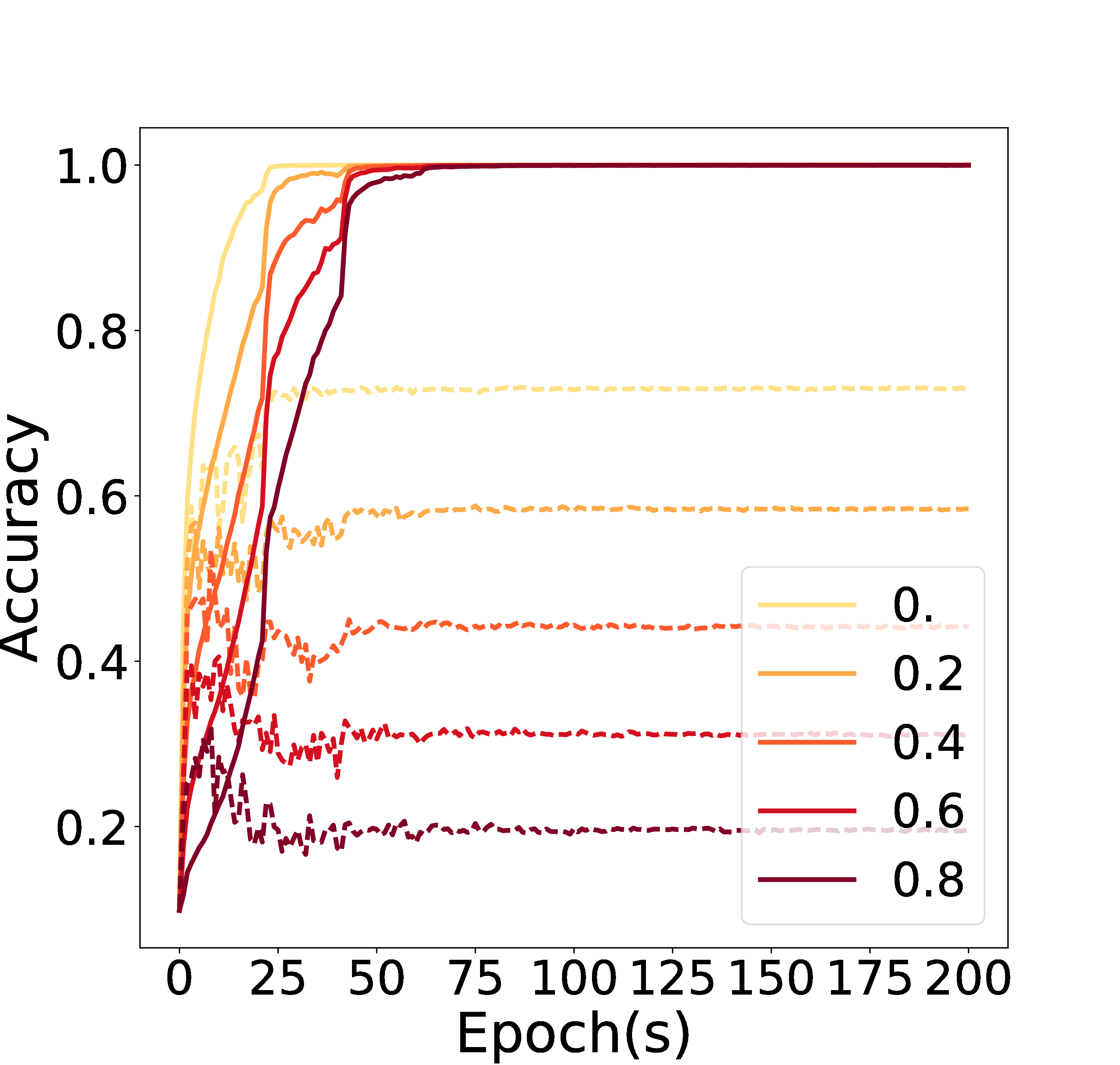}
                   \includegraphics[width=4.1cm]{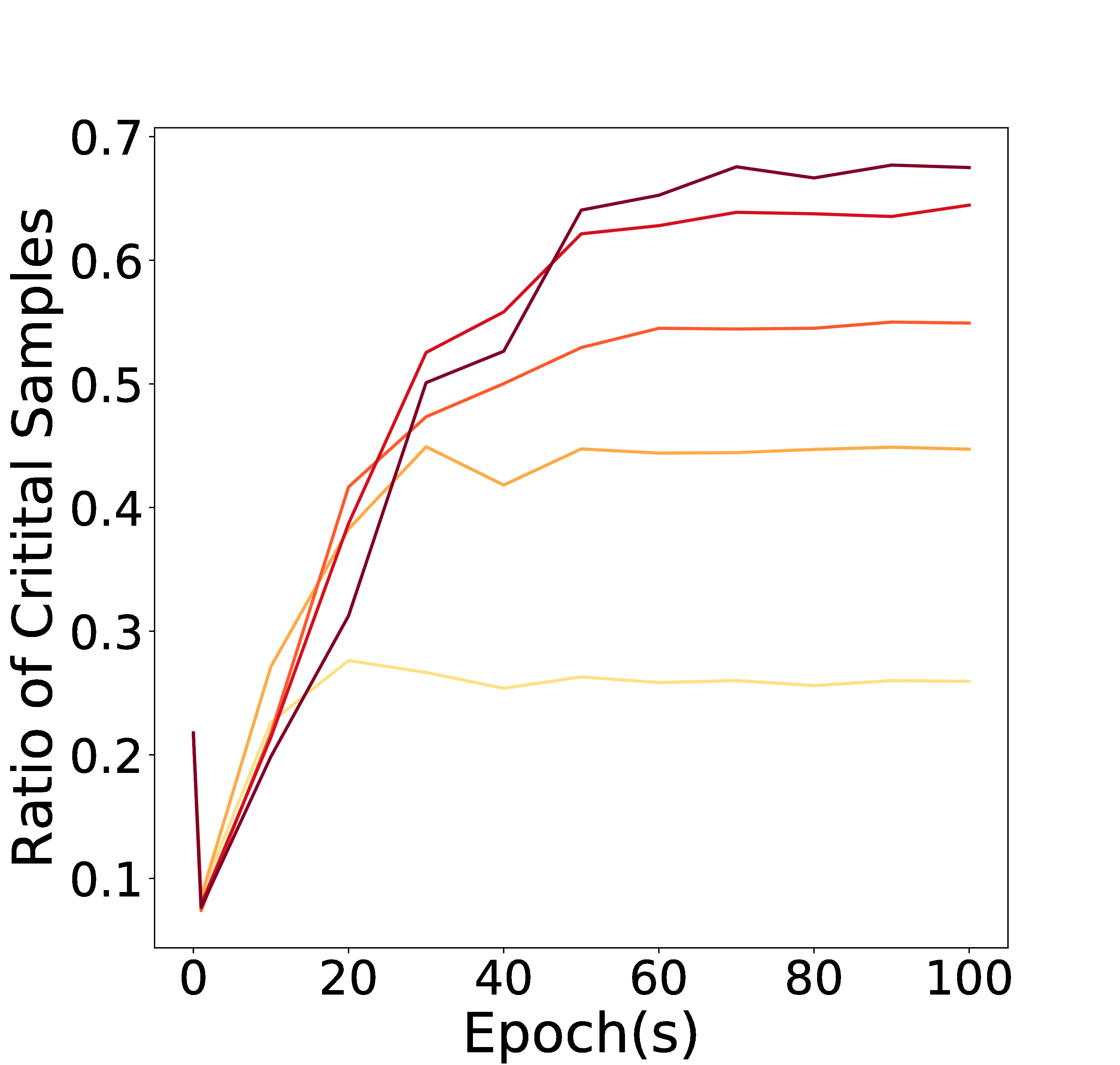}
		   \label{fig:cifar10_noisey_critical}
		}
		\caption{Accuracy (left in each pair, solid is train, dotted is validation) and Critical sample ratios (right in each pair) for CIFAR10.}
	
    \label{fig:cifar_nicolas}
	\end{figure*}

\section{DNNs Learn Patterns First}
\label{sec:simple_patterns_first}
This section aims at studying how the complexity of \af{the hypotheses} learned by DNNs evolve during training for real data vs.~noise data. 
To achieve this goal, we build on the intuition that the number of different decision regions into which an input space is partitioned reflects the complexity of the learned hypothesis~\citep{sokolic2016robust}. 
This notion is similar in spirit to the degree to which a function can scatter random labels: a higher density of decision boundaries in the data space allows more samples to be scattered.

\af{Therefore, we estimate the complexity} by measuring how densely points on the data manifold are present around the model's decision boundaries. 
Intuitively, if we were to randomly sample points from the data distribution, a smaller fraction of points in the proximity of a decision boundary suggests that the learned hypothesis is simpler.

\subsection{Critical Sample Ratio (CSR)}
Here we introduce the notion of a {\it critical sample}, which we use to estimate the density of decision boundaries as discussed above.
Critical samples are a subset of a dataset such that for each such sample $\mathbf{x}$, there exists at least one adversarial example $\hat{\mathbf{x}}$ in the proximity of $\mathbf{x}$. 
Specifically, consider a classification network's output vector $ f(\mathbf{x}) =(f_1(\mathbf{x}), \dots, f_k(\mathbf{x})) \in \mathbb{R}^k $ for a given input sample $ \mathbf{x} \in \mathbb{R}^{n}$ from the data manifold. 
Formally we call a dataset sample $ \mathbf{x} $ a \emph{critical sample} if there exists a point $ \hat{\mathbf{x}} $ such that,
\begin{align}
\label{eq_sens}
&\arg \max_{i} f_{i}(\mathbf{x}) \neq \arg \max_{j} f_{j}(\hat{\mathbf{x}}) \mspace{5mu}  \\ \nonumber
& \mbox{s.t.  } \lVert \mathbf{x} - \hat{\mathbf{x}} \rVert_{\infty} \leq r
\end{align}
where $r$ is a fixed box size.
As in recent work on adversarial examples \citep{kurakin2016adversarial} the above definition depends only on the predicted label $ \arg \max_{i} f_{i}(\mathbf{x}) $ of $ \mathbf{x} $, and not the true label (as in earlier work on adversarial examples, such as \citet{adversarial_examples,goodfellow2014explaining}).

\af{Following the above argument relating complexity to decision boundaries}, a higher number of critical samples indicates a more complex hypothesis. Thus, we measure complexity as the {\it critical sample ratio (CSR)}, that is, the fraction of data-points in a set $|\mathcal{D}|$ for which we can find a critical sample: $ \frac{\# \text{critical samples}}{|\mathcal{D}|}$.

To identify whether a given data point $ \mathbf{x} $ is a critical samples, we search for an adversarial sample $ \hat{\mathbf{x}} $ within a box of radius $r$.
To perform this search, we propose using Langevin dynamics applied to the fast gradient sign method (FGSM, \citet{goodfellow2014explaining}) as shown in algorithm \ref{algo_sens_search}\footnote{In our experiments, we set $\alpha=0.25$, $\beta=0.2$ and $\eta$ is samples from standard normal distribution.}.
We refer to this method as Langevin adversarial sample search (LASS).
While the FGSM search algorithm can get stuck at a points with zero gradient, LASS explores the box more thoroughly.
Specifically, a problem with first order gradient search methods (like FGSM) is that there might exist training points where the gradient is 0, but with a large $2^{nd}$ derivative corresponding to a large change in prediction in the neighborhood. 
The noise added by the LASS algorithm during the search enables escaping from such points.

\begin{algorithm}[!ht]
	\caption{Langevin Adversarial Sample Search (LASS)}
	\begin{algorithmic}[1]
		\REQUIRE $\mathbf{x} \in \mathbb{R}^{n}$, $\alpha$, $\beta$, $ r $, noise process $ \eta $
		\ENSURE $\hat{\mathbf{x}}$
        \STATE converged = FALSE
		\STATE $ \tilde{\mathbf{x}} \leftarrow \mathbf{x} $; $ \hat{\mathbf{x}} \leftarrow \emptyset$
		\WHILE {not converged or max iter reached}
		\STATE $\Delta = \alpha \cdot \mbox{sign}(\frac{\partial f_{k}(\mathbf{x})}{\partial \mathbf{x}}) + \beta \cdot \eta$
		\STATE $ \tilde{\mathbf{x}}  \leftarrow \tilde{\mathbf{x}} + \Delta$
		\FOR { $ i \in [n]$}
		\STATE  $ \tilde{\mathbf{x}}_{i}  \leftarrow \left\{ \begin{array}{ll}
		{\mathbf{x}}_{i} + r \cdot \mbox{sign}(\tilde{\mathbf{x}}_{i}- {\mathbf{x}}^{i}) & if \lvert \tilde{\mathbf{x}}_{i}- {\mathbf{x}}_{i} \rvert > r \\
		\tilde{\mathbf{x}}_{i} &  otherwise \\
		\end{array} 
		\right.$
		\ENDFOR
        \IF{$\arg \max_{i} f(\mathbf{x}) \neq \arg \max_{i} f(\tilde{\mathbf{x}})$}
        \STATE converged = TRUE
        \STATE $\hat{\mathbf{x}} \leftarrow \tilde{\mathbf{x}}$
        \ENDIF
		\ENDWHILE
	\end{algorithmic}
    \label{algo_sens_search}
\end{algorithm}

\begin{figure}[!t]
	\center
	\label{fig_valid_stability}
	\includegraphics[width=5cm]{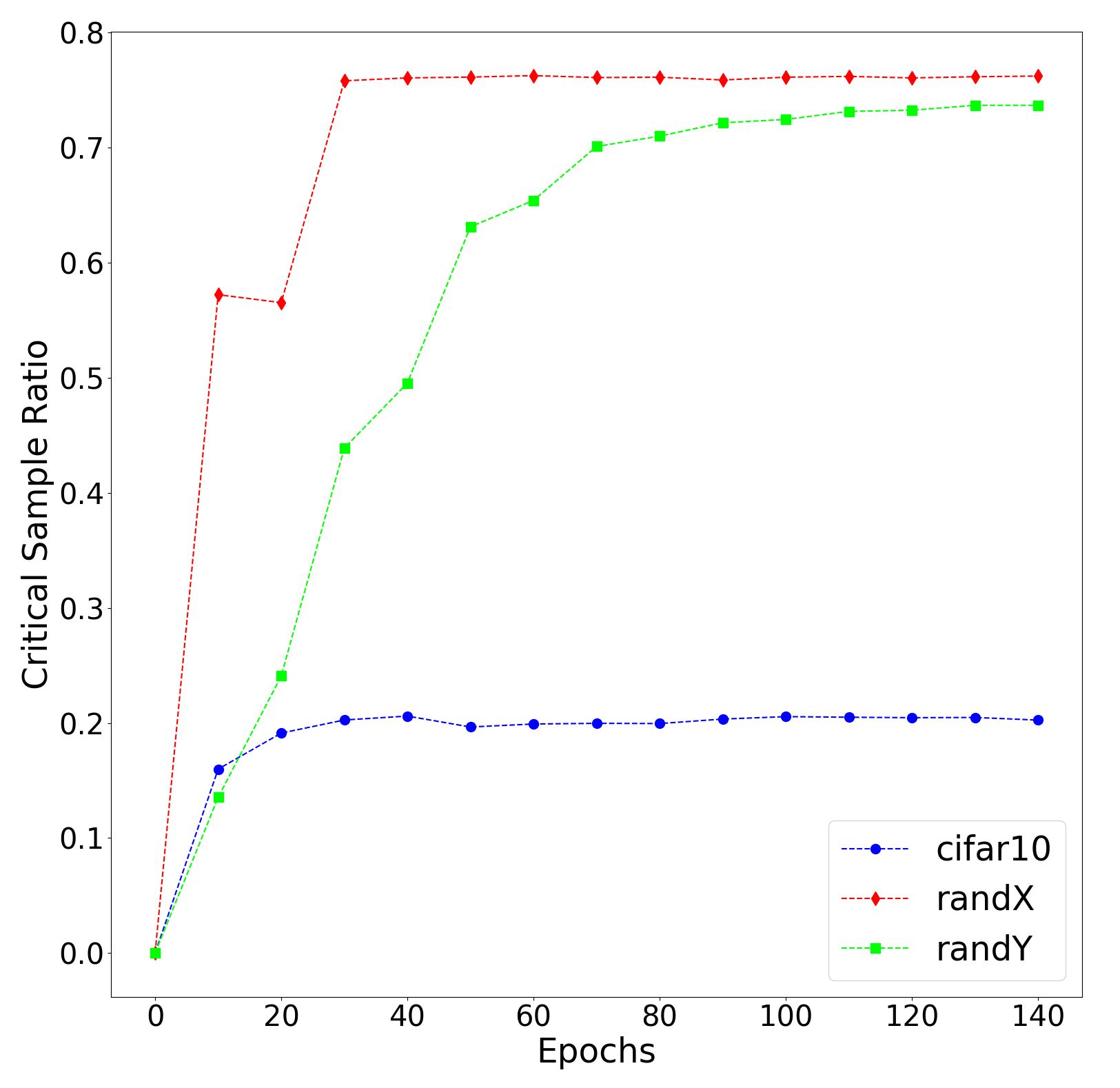}
	\caption{Critical sample ratio throughout training on CIFAR-10, random input (randX), and random label (randY) datasets.}
\end{figure}

\subsection{Critical Samples Throughout Training}
\label{sec_CS}
We now show that the number of critical samples is much higher for a deep network (specifically, a CNN) trained on noise data compared with real data.
To do so, we measure the number of critical samples in the validation set\footnote{
We also measure the number of critical samples in the training sets. Since we train our models using log loss, training points are pushed away from the decision boundary even after the network learns to classify them correctly. This leads to an initial rise and then fall of the number of critical samples in the training sets.
},
throughout training\footnote{We use a box size of 0.3, which is small enough in a 0-255 pixel scale to be unnoticeable by a human evaluator. \af{Different values for $r$ were tested but did not change results qualitatively and lead to the same conclusions}}.
Results are shown in Figure \ref{fig_valid_stability}.
A higher number of critical samples \af{for models trained on} noise data compared with those \af{trained on} real data suggests that the learned decision surface is more complex for noise data (randX and randY).
We also observe that the CSR increases gradually with increasing number of epochs and then stabilizes. This suggests that the networks learn gradually more complex hypotheses during training for all three datasets.

In our next experiment, we evaluate the performance and critical sample ratio of datasets with $ 20\% $ to  $ 80 \% $ of the training data replaced with either input or label noise.
Results for MNIST and CIFAR-10 are shown in Figures \ref{fig:mnist_nicolas} and \ref{fig:cifar_nicolas}, respectively.
For both randX and randY datasets, the CSR is higher for noisier datasets, reflecting the higher level of complexity of the learned prediction function. 
The final and maximum validation accuracies are also both lower for noisier datasets, indicating that the noise examples interfere somewhat with the networks ability to learn about the real data.

More significantly, for randY datasets (Figures \ref{fig:mnist_noisey_critical} and \ref{fig:cifar10_noisey_critical}), the network achieves maximum accuracy on the validation set before achieving high accuracy on the training set.
Thus the model first learns the simple and general patterns of the real data before fitting the noise (which results in decreasing validation accuracy).
Furthermore, as the model moves from fitting real data to fitting noise, the CSR greatly increases, indicating the need for more complex hypotheses to explain the noise.
Combining this result with our results from Section~\ref{sec:easy_examples}, we conclude that real data examples are easier to fit than noise.

\section{Effect of Regularization on Learning}
\label{sec:regularization}
Here we demonstrate the ability of regularization to degrade training performance on data with random labels, while maintaining generalization performance on real data.
\citet{understanding_DL} argue that explicit regularizations are not the main explanation of good generalization performance, rather SGD based optimization is largely responsible for it. Our findings extend their claim and indicate that explicit regularizations can substantially limit the speed of memorization of noise data without significantly impacting learning on real data.

\begin{figure}[!t]
        \center
        \label{fig:reg_pattern}
        \includegraphics[width=5.5cm]{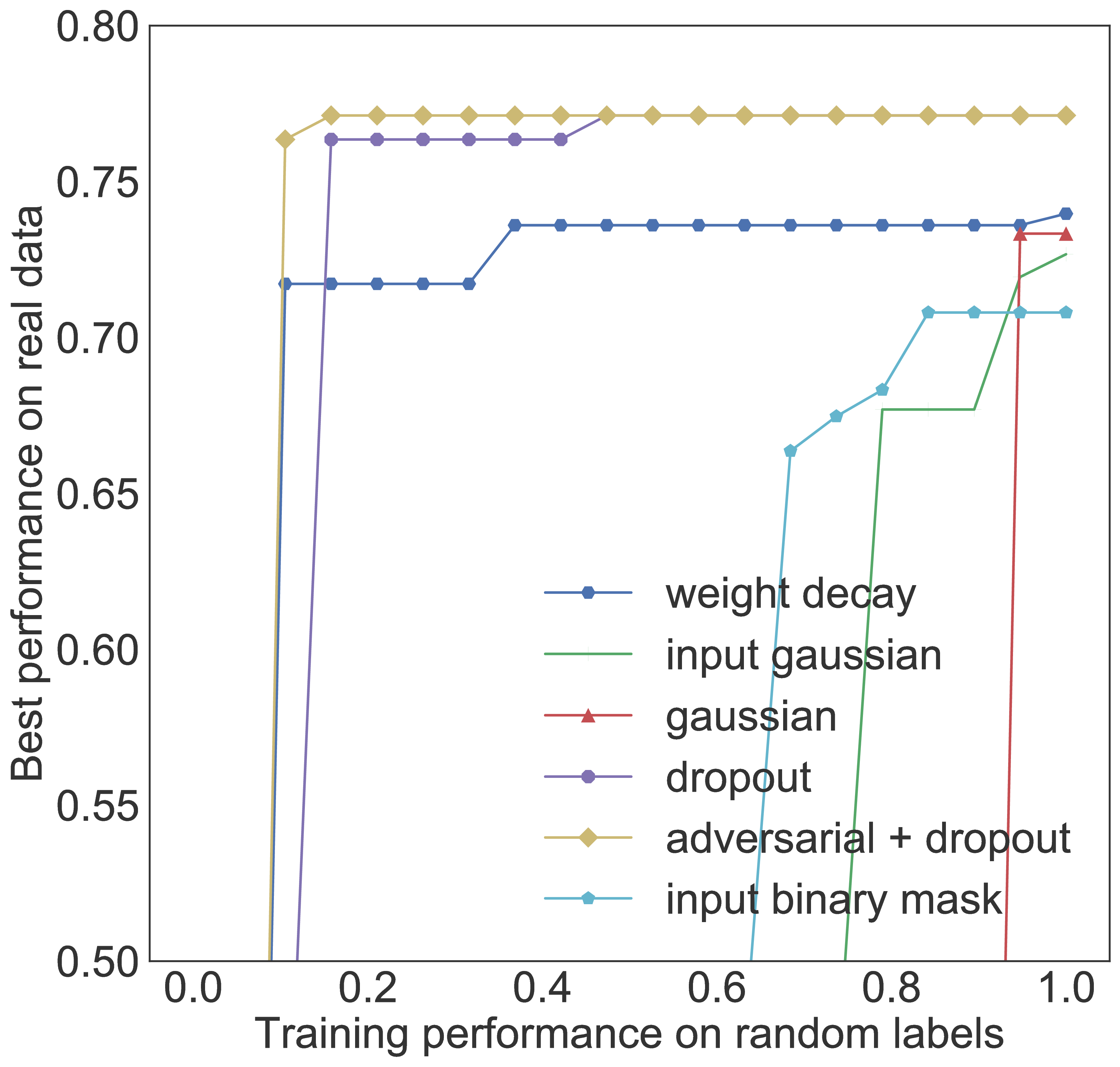}
        \caption{Effect of different regularizers on train accuracy (on noise dataset) vs.~validation accuracy (on real dataset). Flatter curves indicate that memorization (on noise) can be capped without sacrificing generalization (on real data).}
    \end{figure}

\begin{figure}[!t]
        \center
        \label{fig:randY_capping}
        \includegraphics[width=4cm]{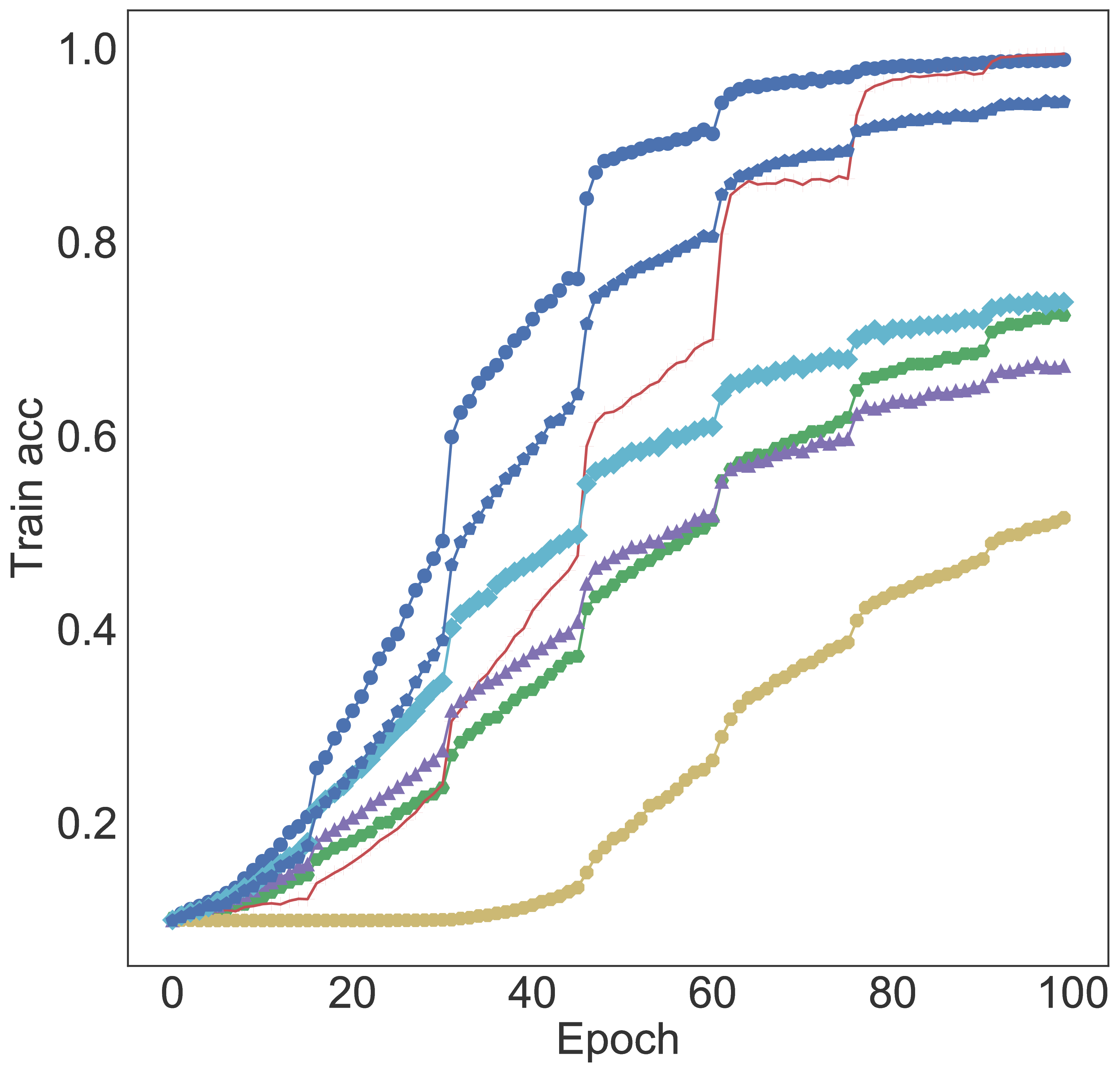}
        \includegraphics[width=3.8cm]{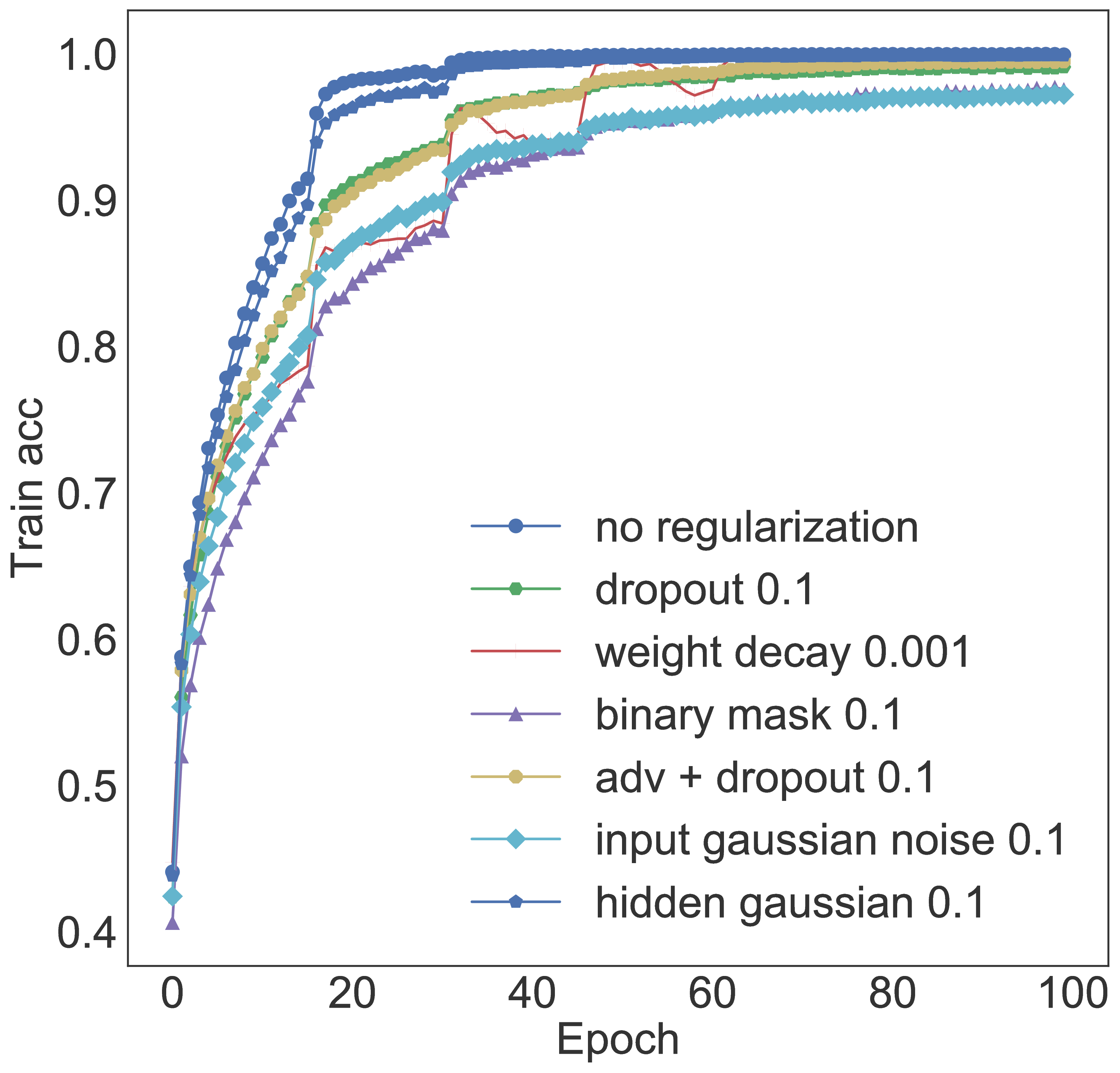}
        \caption{Training curves for different regularization techniques on random label (left) and real (right) data.  The vertical ordering of the curves is different for random labels than for real data, indicating differences in the propensity of different regularizers to slow-down memorization.}
    \end{figure}

{
We compare the performance of CNNs trained on CIFAR-10 and randY with the following regularizers: dropout (with dropout rates in range  $0$-$0.9$), input dropout (range $0$-$0.9$),  input Gaussian noise (with standard deviation in range $0$-$5$), hidden Gaussian noise (range $0$-$0.3$), weight decay (range $0$-$1$) and additionally dropout with adversarial training (with weighting factor in range $0.2$-$0.7$ and dropout in rate range $0.03$-$0.5$).\footnote{We perform adversarial training using critical samples found by LASS algorithm with default parameters. }
We train a separate model for every combination of dataset, regularization technique, and regularization parameter.

\da{The results are summarized in Figure~\ref{fig:reg_pattern}.
\af{For each combination of dataset and regularization technique, the final training accuracy on randY (x-axis) is plotted against the best validation accuracy on CIFAR-10 from amongst the models trained with different regularization parameters (y-axis).} 
Flat curves indicate that the corresponding regularization technique can reduce memorization when applied on random labeling, while resulting in the same validation accuracy \af{on the clean validation set}.}
Our results show that different regularizers target memorization behavior to different extent -- dropout being the most effective. 
We find that dropout, especially coupled with adversarial training, is best at hindering memorization without reducing the model's ability to learn.
Figure~\ref{fig:randY_capping} additionally shows this effect for selected experiments (i.e. selected hyperparameter values) in terms of train loss.}

\section{Related Work}

Our work builds on the experiments and challenges the interpretations of \citet{understanding_DL}.
We make heavy use of their methodology of studying DNN training in the context of noise datasets. \citet{understanding_DL} show that DNNs can perfectly fit noise and thus that their generalization ability cannot be explained through traditional statistical learning theory (e.g., see~\cite{vapnik1998statistical, bartlett2005local}). 
We agree with this finding, but show in addition that the degree of memorization and generalization in DNNs depends not only on the architecture and training procedure (including explicit regularizations), \textit{but also on the training data itself}\footnote{We conclude the latter part based on experimental findings in sections \ref{sec:qualitative_differences} and \ref{sec_CS}}. 

Another direction we investigate is the relationship between regularization and memorization. 
\citet{understanding_DL} argue that explicit and implicit regularizers (including SGD) might not explain or limit shattering of random data. 
In this work we show that regularizers (especially dropout) {\it do} control the \emph{speed} at which DNNs memorize. This is interesting since dropout is also known to prevent catastrophic forgetting \cite{goodfellow2013empirical} and thus in general it seems to help DNNs retain patterns.

A number of arguments support the idea that SGD-based learning imparts a regularization effect, especially with a small batch size~\citep{wilson2003general} or a small number of epochs~\citep{hardt2015train}.
Previous work also suggests that SGD prioritizes the learning of simple hypothesis first. 
\citet{CIS-125961} showed that, for linear models, SGD first learns models with small $\ell^2$ parameter norm.
More generally, the efficacy of early stopping shows that SGD first learns simpler models~\citep{yao2007early}. 
We extend these results, showing that DNNs trained with SGD learn patterns before memorizing, {\it even in the presence of noise examples}. 

Various previous works have analyzed explanations for the generalization power of DNNs.
\citet{Montavon2011} use kernel methods to analyze the complexity of deep learning architectures, and find that network priors (e.g.~implemented by the network structure of a CNN or MLP) control the speed of learning at each layer. 
\citet{neyshabur2014search} note that the number of parameters does not control the effective capacity of a DNN, and that the reason for DNNs' generalization is unknown. 
We supplement this result by showing how the impact of representational capacity changes with varying noise levels. 
While exploring the effect of noise samples on learning dynamics has a long tradition~\citep{bishop1995training, an1996effects}, we are the first to examine {\it relationships} between the fraction of noise samples and other attributes of the learning algorithm, namely: capacity, training time and dataset size.

Multiple techniques for analyzing the training of DNNs have been proposed before, including looking at generalization error, trajectory length evolution~\citep{raghu2016expressive}, analyzing Jacobians associated to different layers~\citep{wang2016analysis, saxe2013exact}, or the shape of the loss minima found by SGD~\citep{im2016empirical, chaudhari2016entropy, keskar2016large}.
Instead of measuring the sharpness of the loss for the learned hypothesis, we investigate the complexity of the learned hypothesis throughout training and across different datasets and regularizers, as measured by the critical sample ratio. 
Critical samples refer to real data-points that have adversarial examples \citep{adversarial_examples,goodfellow2014explaining} nearby.
Adversarial examples originally referred to imperceptibly perturbed data-points that are confidently misclassified.
\citep{ miyato2015distributional} define {\it virtual} adversarial examples via changes in the predictive distribution instead, thus extending the definition to unlabeled data-points.
\citet{kurakin2016adversarial} recommend using this definition when training on adversarial examples, and it is the definition we use.

Two contemporary works perform in-depth explorations of topics related to our work.
\citet{bojanowski2017} show that predicting random noise targets can yield state of the art results in unsupervised learning, corroborating our findings in Section ~\ref{sec:easy_examples}, especially Figure ~\ref{fig:filters.pdf}.
\citet{koh2017understanding} use {\it influence functions} to measure the impact on parameter changes during training, as in our Section~\ref{sec:grad_sens}.
They explore several promising applications for this technique, including generation of adversarial {\it training} examples.

\section{Conclusion}

Our empirical exploration demonstrates qualitative differences in DNN optimization on noise vs.~real data, all of which support the claim that DNNs trained with SGD-variants first use patterns, not brute force memorization, to fit real data. 
However, since DNNs have the demonstrated ability to fit noise, it is unclear why they find generalizable solutions on real data; we believe that the deep learning priors including distributed and hierarchical representations likely play an important role.
Our analysis suggests that memorization and generalization in DNNs depend on network architecture and optimization procedure, but also on the data itself.
We hope to encourage future research on how properties of datasets influence the behavior of deep learning algorithms, and suggest a data-dependent understanding of DNN capacity as a research goal.

\subsubsection*{Acknowledgments}
We thank Akram Erraqabi, Jason Jo and Ian Goodfellow for helpful discussions. 
SJ was supported by Grant No.~DI 2014/016644 from Ministry of Science and Higher Education, Poland. DA was supported by IVADO, CIFAR and NSERC.
EB was financially supported by the Samsung Advanced Institute of Technology (SAIT).
MSK and SJ were supported by MILA during the course of this work.
We acknowledge the computing resources provided by ComputeCanada and CalculQuebec.
Experiments were carried out using Theano~\citep{theano} and Keras~\citep{keras}. 
    
\bibliography{main}

\begin{thebibliography}{40}
\providecommand{\natexlab}[1]{#1}
\providecommand{\url}[1]{\texttt{#1}}
\expandafter\ifx\csname urlstyle\endcsname\relax
  \providecommand{\doi}[1]{doi: #1}\else
  \providecommand{\doi}{doi: \begingroup \urlstyle{rm}\Url}\fi

\bibitem[An(1996)]{an1996effects}
An, Guozhong.
\newblock The effects of adding noise during backpropagation training on a
  generalization performance.
\newblock \emph{Neural computation}, 8\penalty0 (3):\penalty0 643--674, 1996.

\bibitem[Bartlett et~al.(2005)Bartlett, Bousquet, Mendelson,
  et~al.]{bartlett2005local}
Bartlett, Peter~L, Bousquet, Olivier, Mendelson, Shahar, et~al.
\newblock Local rademacher complexities.
\newblock \emph{The Annals of Statistics}, 33\penalty0 (4):\penalty0
  1497--1537, 2005.

\bibitem[Bengio et~al.(2009)]{bengio2009}
Bengio, Yoshua et~al.
\newblock Learning deep architectures for ai.
\newblock \emph{Foundations and trends{\textregistered} in Machine Learning},
  2\penalty0 (1):\penalty0 1--127, 2009.

\bibitem[Bishop(1995)]{bishop1995training}
Bishop, Chris~M.
\newblock Training with noise is equivalent to tikhonov regularization.
\newblock \emph{Neural computation}, 7\penalty0 (1):\penalty0 108--116, 1995.

\bibitem[{Bojanowski} \& {Joulin}(2017){Bojanowski} and
  {Joulin}]{bojanowski2017}
{Bojanowski}, P. and {Joulin}, A.
\newblock {Unsupervised Learning by Predicting Noise}.
\newblock \emph{ArXiv e-prints}, April 2017.

\bibitem[Bottou(1998)]{bottou1998online}
Bottou, L{\'e}on.
\newblock Online learning and stochastic approximations.
\newblock \emph{On-line learning in neural networks}, 17\penalty0 (9):\penalty0
  142, 1998.

\bibitem[Chaudhari et~al.(2016)Chaudhari, Choromanska, Soatto, and
  LeCun]{chaudhari2016entropy}
Chaudhari, Pratik, Choromanska, Anna, Soatto, Stefano, and LeCun, Yann.
\newblock Entropy-sgd: Biasing gradient descent into wide valleys.
\newblock \emph{arXiv preprint arXiv:1611.01838}, 2016.

\bibitem[Chollet et~al.(2015)]{keras}
Chollet, Fran\c{c}ois et~al.
\newblock Keras.
\newblock \url{https://github.com/fchollet/keras}, 2015.

\bibitem[Cybenko(1989)]{cybenko1989approximation}
Cybenko, George.
\newblock Approximation by superpositions of a sigmoidal function.
\newblock \emph{Mathematics of Control, Signals, and Systems (MCSS)},
  2\penalty0 (4):\penalty0 303--314, 1989.

\bibitem[Fix \& Hodges~Jr(1951)Fix and Hodges~Jr]{fix1951discriminatory}
Fix, Evelyn and Hodges~Jr, Joseph~L.
\newblock Discriminatory analysis-nonparametric discrimination: consistency
  properties.
\newblock Technical report, DTIC Document, 1951.

\bibitem[Gini(1913)]{gini}
Gini, Corrado.
\newblock Variabilita e mutabilita.
\newblock \emph{Journal of the Royal Statistical Society}, 76\penalty0 (3),
  1913.

\bibitem[Goodfellow et~al.(2016)Goodfellow, Bengio, and Courville]{dl_book}
Goodfellow, Ian, Bengio, Yoshua, and Courville, Aaron.
\newblock \emph{Deep Learning}.
\newblock MIT Press, 2016.
\newblock \url{http://www.deeplearningbook.org}.

\bibitem[Goodfellow et~al.(2013)Goodfellow, Mirza, Xiao, Courville, and
  Bengio]{goodfellow2013empirical}
Goodfellow, Ian~J, Mirza, Mehdi, Xiao, Da, Courville, Aaron, and Bengio,
  Yoshua.
\newblock An empirical investigation of catastrophic forgetting in
  gradient-based neural networks.
\newblock \emph{arXiv preprint arXiv:1312.6211}, 2013.

\bibitem[Goodfellow et~al.(2014)Goodfellow, Shlens, and
  Szegedy]{goodfellow2014explaining}
Goodfellow, Ian~J, Shlens, Jonathon, and Szegedy, Christian.
\newblock Explaining and harnessing adversarial examples.
\newblock \emph{arXiv preprint arXiv:1412.6572}, 2014.

\bibitem[Hardt et~al.(2015)Hardt, Recht, and Singer]{hardt2015train}
Hardt, Moritz, Recht, Benjamin, and Singer, Yoram.
\newblock Train faster, generalize better: Stability of stochastic gradient
  descent.
\newblock \emph{arXiv preprint arXiv:1509.01240}, 2015.

\bibitem[Hornik et~al.(1989)Hornik, Stinchcombe, and
  White]{hornik1989multilayer}
Hornik, Kurt, Stinchcombe, Maxwell, and White, Halbert.
\newblock Multilayer feedforward networks are universal approximators.
\newblock \emph{Neural networks}, 2\penalty0 (5):\penalty0 359--366, 1989.

\bibitem[Im et~al.(2016)Im, Tao, and Branson]{im2016empirical}
Im, Daniel~Jiwoong, Tao, Michael, and Branson, Kristin.
\newblock An empirical analysis of deep network loss surfaces.
\newblock \emph{arXiv preprint arXiv:1612.04010}, 2016.

\bibitem[Keskar et~al.(2016)Keskar, Mudigere, Nocedal, Smelyanskiy, and
  Tang]{keskar2016large}
Keskar, Nitish~Shirish, Mudigere, Dheevatsa, Nocedal, Jorge, Smelyanskiy,
  Mikhail, and Tang, Ping Tak~Peter.
\newblock On large-batch training for deep learning: Generalization gap and
  sharp minima.
\newblock \emph{arXiv preprint arXiv:1609.04836}, 2016.

\bibitem[Koh \& Liang(2017)Koh and Liang]{koh2017understanding}
Koh, Pang~Wei and Liang, Percy.
\newblock Understanding black-box predictions via influence functions.
\newblock \emph{arXiv preprint arXiv:1703.04730}, 2017.

\bibitem[Krizhevsky et~al.()Krizhevsky, Nair, and Hinton]{CIFAR10}
Krizhevsky, Alex, Nair, Vinod, and Hinton, Geoffrey.
\newblock Cifar-10 (canadian institute for advanced research).
\newblock URL \url{http://www.cs.toronto.edu/~kriz/cifar.html}.

\bibitem[Kurakin et~al.(2016)Kurakin, Goodfellow, and
  Bengio]{kurakin2016adversarial}
Kurakin, Alexey, Goodfellow, Ian, and Bengio, Samy.
\newblock Adversarial examples in the physical world.
\newblock \emph{arXiv preprint arXiv:1607.02533}, 2016.

\bibitem[LeCun et~al.(1998)LeCun, Cortes, and Burges]{lecun1998mnist}
LeCun, Yann, Cortes, Corinna, and Burges, Christopher~JC.
\newblock The mnist database of handwritten digits, 1998.

\bibitem[Lin \& Tegmark(2016)Lin and Tegmark]{Tegmark}
Lin, Henry~W and Tegmark, Max.
\newblock Why does deep and cheap learning work so well?
\newblock \emph{arXiv preprint arXiv:1608.08225}, 2016.

\bibitem[Maclaurin et~al.(2015)Maclaurin, Duvenaud, and
  Adams]{maclaurin2015gradient}
Maclaurin, Dougal, Duvenaud, David~K, and Adams, Ryan~P.
\newblock Gradient-based hyperparameter optimization through reversible
  learning.
\newblock In \emph{ICML}, pp.\  2113--2122, 2015.

\bibitem[Miyato et~al.(2015)Miyato, Maeda, Koyama, Nakae, and
  Ishii]{miyato2015distributional}
Miyato, Takeru, Maeda, Shin-ichi, Koyama, Masanori, Nakae, Ken, and Ishii,
  Shin.
\newblock Distributional smoothing with virtual adversarial training.
\newblock \emph{stat}, 1050:\penalty0 25, 2015.

\bibitem[Montavon et~al.(2011)Montavon, Braun, and M\"{u}ller]{Montavon2011}
Montavon, Gr{\'e}goire, Braun, Mikio~L., and M\"{u}ller, Klaus-Robert.
\newblock Kernel analysis of deep networks.
\newblock \emph{Journal of Machine Learning Research}, 12, 2011.

\bibitem[Montufar et~al.(2014)Montufar, Pascanu, Cho, and Bengio]{montufar2014}
Montufar, Guido~F, Pascanu, Razvan, Cho, Kyunghyun, and Bengio, Yoshua.
\newblock On the number of linear regions of deep neural networks.
\newblock In Ghahramani, Z., Welling, M., Cortes, C., Lawrence, N.~D., and
  Weinberger, K.~Q. (eds.), \emph{Advances in Neural Information Processing
  Systems 27}, pp.\  2924--2932. Curran Associates, Inc., 2014.

\bibitem[Neyshabur et~al.(2014)Neyshabur, Tomioka, and
  Srebro]{neyshabur2014search}
Neyshabur, Behnam, Tomioka, Ryota, and Srebro, Nathan.
\newblock In search of the real inductive bias: On the role of implicit
  regularization in deep learning.
\newblock \emph{arXiv preprint arXiv:1412.6614}, 2014.

\bibitem[Poole et~al.(2016)Poole, Lahiri, Raghu, Sohl-Dickstein, and
  Ganguli]{ganguli2016}
Poole, Ben, Lahiri, Subhaneil, Raghu, Maithreyi, Sohl-Dickstein, Jascha, and
  Ganguli, Surya.
\newblock Exponential expressivity in deep neural networks through transient
  chaos.
\newblock In Lee, D.~D., Sugiyama, M., Luxburg, U.~V., Guyon, I., and Garnett,
  R. (eds.), \emph{Advances in Neural Information Processing Systems 29}, pp.\
  3360--3368. Curran Associates, Inc., 2016.

\bibitem[Raghu et~al.(2016)Raghu, Poole, Kleinberg, Ganguli, and
  Sohl-Dickstein]{raghu2016expressive}
Raghu, Maithra, Poole, Ben, Kleinberg, Jon, Ganguli, Surya, and Sohl-Dickstein,
  Jascha.
\newblock On the expressive power of deep neural networks.
\newblock \emph{arXiv preprint arXiv:1606.05336}, 2016.

\bibitem[Saxe et~al.(2013)Saxe, McClelland, and Ganguli]{saxe2013exact}
Saxe, Andrew~M, McClelland, James~L, and Ganguli, Surya.
\newblock Exact solutions to the nonlinear dynamics of learning in deep linear
  neural networks.
\newblock \emph{arXiv preprint arXiv:1312.6120}, 2013.

\bibitem[Sjoberg et~al.(1995)Sjoberg, Sjoeberg, Sjöberg, and
  Ljung]{CIS-125961}
Sjoberg, J., Sjoeberg, J., Sjöberg, J., and Ljung, L.
\newblock Overtraining, regularization and searching for a minimum, with
  application to neural networks.
\newblock \emph{International Journal of Control}, 62:\penalty0 1391--1407,
  1995.

\bibitem[Sokolic et~al.(2016)Sokolic, Giryes, Sapiro, and
  Rodrigues]{sokolic2016robust}
Sokolic, Jure, Giryes, Raja, Sapiro, Guillermo, and Rodrigues, Miguel~RD.
\newblock Robust large margin deep neural networks.
\newblock \emph{arXiv preprint arXiv:1605.08254}, 2016.

\bibitem[Szegedy et~al.(2013)Szegedy, Zaremba, Sutskever, Bruna, Erhan,
  Goodfellow, and Fergus]{adversarial_examples}
Szegedy, Christian, Zaremba, Wojciech, Sutskever, Ilya, Bruna, Joan, Erhan,
  Dumitru, Goodfellow, Ian~J., and Fergus, Rob.
\newblock Intriguing properties of neural networks.
\newblock \emph{CoRR}, abs/1312.6199, 2013.
\newblock URL \url{http://arxiv.org/abs/1312.6199}.

\bibitem[Theano Development~Team(2016)]{theano}
Theano Development~Team, {and others}.
\newblock {Theano: A {Python} framework for fast computation of mathematical
  expressions}.
\newblock \emph{arXiv e-prints}, abs/1605.02688, May 2016.

\bibitem[Vapnik \& Vapnik(1998)Vapnik and Vapnik]{vapnik1998statistical}
Vapnik, Vladimir~Naumovich and Vapnik, Vlamimir.
\newblock \emph{Statistical learning theory}, volume~1.
\newblock Wiley New York, 1998.

\bibitem[Wang()]{wang2016analysis}
Wang, Shengjie.
\newblock Analysis of deep neural networks with the extended data jacobian
  matrix.

\bibitem[Wilson \& Martinez(2003)Wilson and Martinez]{wilson2003general}
Wilson, D~Randall and Martinez, Tony~R.
\newblock The general inefficiency of batch training for gradient descent
  learning.
\newblock \emph{Neural Networks}, 16\penalty0 (10):\penalty0 1429--1451, 2003.

\bibitem[Yao et~al.(2007)Yao, Rosasco, and Caponnetto]{yao2007early}
Yao, Yuan, Rosasco, Lorenzo, and Caponnetto, Andrea.
\newblock On early stopping in gradient descent learning.
\newblock \emph{Constructive Approximation}, 26\penalty0 (2):\penalty0
  289--315, 2007.

\bibitem[Zhang et~al.(2017)Zhang, Bengio, Hardt, Recht, and
  Vinyals]{understanding_DL}
Zhang, Chiyuan, Bengio, Samy, Hardt, Moritz, Recht, Benjamin, and Vinyals,
  Oriol.
\newblock Understanding deep learning requires rethinking generalization.
\newblock \emph{International Conference on Learning Representations (ICLR)},
  2017.

\end{thebibliography}
\bibliographystyle{icml2017}
\end{document}